\documentclass[conference]{IEEEtran}
\IEEEoverridecommandlockouts
\usepackage{amsmath,amssymb,amsfonts}
\usepackage{graphicx, caption, subcaption}
\usepackage{textcomp}
\usepackage{xcolor}

\usepackage{pgfplots}
\usepackage{pgf,tikz}
\usepackage{mathrsfs}
\pgfplotsset{compat=1.5}

\usepackage{color}
\usepackage{xspace}
\usepackage{multirow}

\usepackage[pagebackref=true,breaklinks=true,letterpaper=true,colorlinks,bookmarks=false]{hyperref}

\newcommand{\figref}[1]{\mbox{Fig.~\ref{#1}}}
\newcommand{\tabref}[1]{\mbox{Table~\ref{#1}}}

\definecolor{colorRect2}{RGB}{0,0,128}
\definecolor{colorRect2Agnostic}{RGB}{0,14,255}
\definecolor{colorRect1}{RGB}{0,254,222}
\definecolor{colorRect1Agnostic}{RGB}{45,236,0}

\definecolor{lime}{HTML}{A6CE39}

\def\addlegendimage{\csname pgfplots@addlegendimage\endcsname}
\def\addlegendentry{\csname pgfplots@addlegendentry\endcsname}

\def\ie{\emph{i.e}\onedot}

\def\etal{\emph{et al}\onedot}
\makeatother

\DeclareCaptionLabelFormat{table}{\tablename~\thetable}

\makeatletter

\DeclareRobustCommand\onedot{\futurelet\@let@token\@onedot}
\def\@onedot{\ifx\@let@token.\else.\null\fi\xspace}
\def\etal{\emph{et al}\onedot}

\makeatother

\DeclareRobustCommand{\orcidicon}{
  \begin{tikzpicture}
  \draw[lime, fill=lime] (0,0) 
  circle [radius=0.16] 
  node[white] {{\fontfamily{qag}\selectfont \tiny ID}};
  \draw[white, fill=white] (-0.0625,0.095) 
  circle [radius=0.007];
  \end{tikzpicture}
  \hspace{-2mm}
}

\foreach \x in {A, ..., Z}{%
  \expandafter\xdef\csname orcid\x\endcsname{\noexpand\href{https://orcid.org/\csname orcidauthor\x\endcsname}{\noexpand\orcidicon}}
}

\def\BibTeX{{\rm B\kern-.05em{\sc i\kern-.025em b}\kern-.08em
    T\kern-.1667em\lower.7ex\hbox{E}\kern-.125emX}}
\begin{document}

\title{Oriented Boxes for Accurate Instance Segmentation}

\author{\IEEEauthorblockN{\hspace{0.5cm}Patrick Follmann\orcidA{}}
\IEEEauthorblockA{\textit{Research}\\
\textit{MVTec Software GmbH} \\
Munich, Germany}
\and
\IEEEauthorblockN{\hspace{0.5cm}Rebecca K\"onig\orcidB{}}
\IEEEauthorblockA{\textit{Research}\\
\textit{MVTec Software GmbH} \\
Munich, Germany}
}

\maketitle

\begin{abstract}
State-of-the-art instance-aware semantic segmentation algorithms use axis-aligned bounding boxes as an intermediate processing step to infer the final instance mask output. This often leads to coarse and inaccurate mask proposals due to the following reasons: Axis-aligned boxes have a high background to foreground pixel-ratio, there is a strong variation of mask targets with respect to the underlying box, and neighboring instances frequently reach into the axis-aligned bounding box of the instance mask of interest.

In this work, we overcome these problems by proposing to use oriented boxes as the basis to infer instance masks. We show that oriented instance segmentation improves the mask predictions, especially when objects are diagonally aligned, touching, or overlapping each other.
We evaluate our model on the D2S and Screws datasets and show that we can significantly improve the mask accuracy by 10\% and 12\% mAP compared to instance segmentation using axis-aligned bounding boxes, respectively. On the newly introduced Pill Bags dataset we outperform the baseline using only 10\% of the mask annotations.
\end{abstract}

\begin{IEEEkeywords}
instance segmentation, oriented box detection
\end{IEEEkeywords}

%%%%%%%%% BODY TEXT
\section{Introduction}

The precise localization of objects in natural images is fundamental for many industrial tasks, such as bin-picking or object counting. The detection is often done using axis-aligned boxes \cite{lin2017focal,ren_2015_faster}. However, if objects are deformable, articulated, or diagonally oriented, axis-aligned bounding boxes are often only a very coarse approximation of the objects' locations. Instance-aware semantic segmentation (instance segmentation), as introduced in \cite{lin_2014_coco}, tries to overcome this limitation by predicting a pixel-precise mask for each of the instances.
However, recent instance segmentation methods that are at the top of instance segmentation challenge leaderboards, such as FCIS \cite{li_2016_fully} or variants of Mask RCNN \cite{he_2017_mask}, rely on axis-aligned box proposals to infer the instance mask per box.

%%%%%%%%%%%%%%%%% Intro Figure %%%%%%%%%%%%%%%%%%%
\begin{figure}[t]
  \centering
  \setlength{\tabcolsep}{3pt}
  \begin{tabular}{c c}
    axis-aligned &oriented \\
    \includegraphics[trim={0.35cm 1.1cm 1cm 0},clip,width=0.22\textwidth]{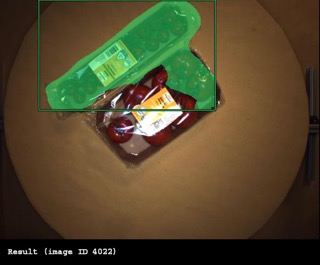}&
    \includegraphics[trim={0.35cm 1.1cm 1cm 0},clip,width=0.22\textwidth]{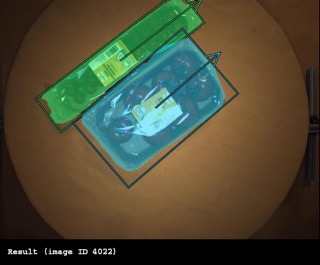}\\
    
    \includegraphics[trim={0.48cm 0.7cm 0.2cm 0},clip,width=0.22\textwidth]{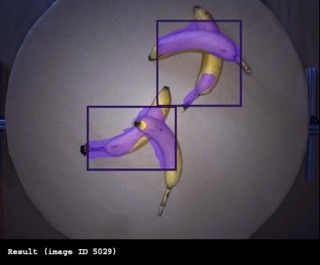}&
    \includegraphics[trim={0.48cm 0.7cm 0.2cm 0},clip,width=0.22\textwidth]{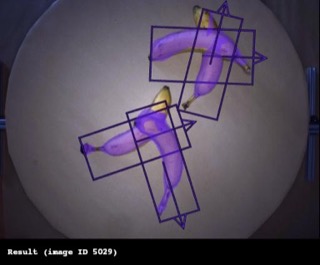}
  \end{tabular}
  \caption{\small{ {\bf Benefits of oriented instance segmentation.} Qualitative comparison of our proposed method based on oriented boxes and the baseline based on axis-aligned boxes. Oriented boxes contain fewer background pixels and avoid reaching into neighboring instances. Resulting instance masks are much more accurate.}}
  \label{fig:IntroImg}
\end{figure}
%%%%%%%%%%%%%%%%% Intro Figure %%%%%%%%%%%%%%%%%%%

This intermediate axis-aligned box detection step introduces several limitations to the final instance mask output:

Depending on an object's orientation, a majority of the box covers the background or another instance that is not of interest for the mask prediction. As features are pooled with respect to the box, this can lead to false classifications or mask predictions reaching into neighboring objects as shown in \figref{fig:IntroImg}. Overlapping bounding boxes also increase the likelihood that one of the candidate boxes is mistakenly filtered out by non-maximum-suppression (NMS).

If an object is rotated, the bounding box aspect-ratio can vary significantly. For unsymmetric objects this leads to highly varying mask targets with respect to the bounding box, even for very precise box predictions which are generally not given. As a consequence, more pixel-wise annotated instances are necessary to learn these variations.

For large and diagonally oriented instances, the relatively coarse resolution often used for mask prediction can lead to inaccurate mask boundaries because of interpolation artifacts. The use of finer grids can resolve this issue to some extent, but comes at the cost of higher computation time and memory consumption.

Therefore, in this work we propose using oriented as opposed to axis-aligned bounding boxes to infer instance masks.
The best possible intersection over union (IoU) an oriented bounding box can achieve for an arbitrary mask is typically much higher than the IoU of the axis-aligned bounding box \cite{bottger_2017_measuring}. Moreover, the oriented instance segmentation (\emph{OIS}) solves the aforementioned issues of axis-aligned instance segmentation (\emph{AAIS}):
\begin{itemize}
  \item Independent of an object's orientation, most of the bounding box area overlaps with the instance of interest. This increases the mask-to-background ratio for mask targets and avoids large overlaps with neighboring objects.
  \item Bounding box aspect ratios become invariant to an object's rotation and the mask exhibits significantly smaller variations with respect to the pooling grid. This leads to more consistent mask targets and a better conditioned training.
  \item For objects with non-overlapping masks, oriented bounding boxes overlap significantly less than axis-aligned bounding boxes. This prevents false positive mask predictions within neighboring objects. 
  %This avoids erroneously filtering out candidate boxes by NMS and makes it easier to tune NMS hyperparameters.
  \item Especially for large, elongated and diagonally aligned objects the accuracy of mask predictions at their boundaries is increased because long edges are aligned with the oriented box.
\end{itemize}

The main contributions of this paper are: We propose to predict accurate instance masks based on oriented box detections. Our approach can be easily applied to existing models based on axis-aligned boxes to enhance their performance. We describe how to adapt architectures for \emph{OIS} and explain necessary changes to different parts of the model compared to baseline \emph{AAIS} models. Our evaluation on three datasets \emph{D2S} \cite{follmann2018d2s}, \emph{Screws} \cite{ulrich2019comparison} and the newly introduced \emph{Pill Bags} dataset shows that the mask accuracy is improved significantly. This leads to a strong increase in overall mAP from 45\% to 55\% on \emph{D2S} with a Mask RCNN architecture \cite{he_2017_mask} and from 46\% to 55\% with a RetinaMask architecture \cite{fu2019retinamask}. On \emph{Screws} the overall mAP is improved from 41\% to 53\%. Moreover, we show that on \emph{Screws} the predicted mask output of our model can be directly used to further refine the oriented box output and improve the box mAP by 1.5\%. %This is in line with the results on \emph{D2S}, where the smallest axis-aligned boxes of the predicted masks of our model outperforms the baseline that was specifically trained for this task. 
Further, we show on \emph{Pill Bags} that for \emph{OIS} only 10\% of the mask annotations are required to outperform the \emph{AAIS} baseline. %\figref{fig:IntroImg} shows some of the mentioned benefits.

%------------------------------------------------------------------------
\section{Related Work}

\paragraph{Instance segmentation.} To date, most instance segmentation methods are based on a Faster RCNN \cite{ren_2015_faster} two-stage object detector. The first stage, \ie the region proposal network (RPN) stage, proposes axis-aligned class-agnostic boxes at all locations where an object is likely to be found. The second stage pools box-specific features with a region of interest (RoI) pooling and refines them to classify the proposals into one of the target classes or background. A second, parallel branch (sometimes with shared weights) is used to further improve the box coordinates via bounding-box regression. In Mask RCNN, He \etal \cite{he_2017_mask} propose to pool features once again, typically with a grid size of $14\times14$, for a third parallel branch that predicts the instance mask. The architecture of the mask prediction branch is similar to the decoder of a fully convolutional network for semantic segmentation \cite{long_2015_fully}. It consists of a number of intermediate convolutions before a transposed convolution upsamples the features by a factor of 2 in each spatial dimension and finally mask probabilites are predicted using a sigmoid activation.
PA-Net \cite{liu2018path} optimizes the information flow within Mask RCNN by fusing the features of several feature pyramid network (FPN) \cite{lin2017fpn} levels to improve the quality, but at the cost of runtime. Mask Scoring RCNN \cite{huang2019mask} learns to predict the quality of the mask predictions, which improves the mAP by recalibrating the predicted scores such that low quality masks also receive a lower score compared to higher quality masks. Among the more recent methods, YOLACT \cite{bolya2019yolact} improves the speed of instance segmentation by using a linear combination of prototypes for mask prediction and a fast variant of non-maximum-suppression. Shape priors are also used in ShapeMask \cite{kuo2019shapemask}, mainly with the goal to improve the generalizability of the model and to reduce the amount of necessary annotated training data. 
RetinaMask \cite{fu2019retinamask} extends the single-stage box detector RetinaNet \cite{lin2017focal} to instance segmentation by adding a mask branch similar to Mask RCNN. 
In contrast to our method, all of the existing instance segmentation methods rely on axis-aligned bounding boxes.

\paragraph{Oriented box detection.} The idea of oriented box detection has been introduced in the context of scene text detection. Existing methods are all based on Faster RCNN \cite{ren_2015_faster}. Jiang \etal \cite{jiang2017r2cnn} still use an axis-aligned box RPN but infer an oriented final box output by regressing the orientation.
In \cite{ma2018arbitrary}, Ma \etal extend the RPN to use oriented anchors and the RoI pooling to pool features with respect to oriented boxes. Oriented box detection has been demonstrated to perform well on other domains than in OCR. For example, Ding \etal \cite{ding2019learning} have set a new benchmark on the oriented object detection dataset DOTA \cite{xia2018dota}. In contrast to our work and \cite{ma2018arbitrary}, they do not use oriented anchors but propose a RoI Transformer module before pooling with respect to the oriented box. Another application that benefits from the use of oriented boxes is ship detection in satellite images \cite{yang2018automatic}, where the authors propose a dense FPN with oriented anchors.

\paragraph{Oriented instance segmentation} To the best of our knowledge, we are the first to combine the approach of \emph{AAIS} methods with oriented box detection to increase the instance mask accuracy. We exemplarily extend the \emph{AAIS} architectures RetinaMask \cite{fu2019retinamask} and Mask RCNN \cite{he_2017_mask} to \emph{OIS}.

%, but replace the axis-aligned box RPN by an oriented box RPN as in \cite{ma2018arbitrary}. We incorporate the idea of \cite{ding2019learning} to pool features with respect to the oriented box proposals. %This ensures that less features are pooled from the background and neighboring objects.
%In comparison to \cite{he_2017_mask}, we use the final boxes instead of the RPN boxes for pooling the mask features, as for oriented boxes it is crucial that the mask prediction fits to the final output box orientation.
%Moreover, none of the previous works has combined oriented box detection with the prediction of instance masks.

%In comparison to previous oriented object detection methods, we explicitly allow to set \emph{classes without orientation}: Since in most applications not all classes have a well-defined orientation we include the option that the model falls back to an axis-aligned box for some user-set classes, which stabilizes the training.

%------------------------------------------------------------------------
\section{Oriented Instance Segmentation}

%%%%%%%%%%%%%%%%% Architecture Figure %%%%%%%%%%%%%%%%%%%
\begin{figure*}[t]
  \centering
  \includegraphics[trim={0 3.2cm 0 2.1cm},clip,width=0.95\textwidth]{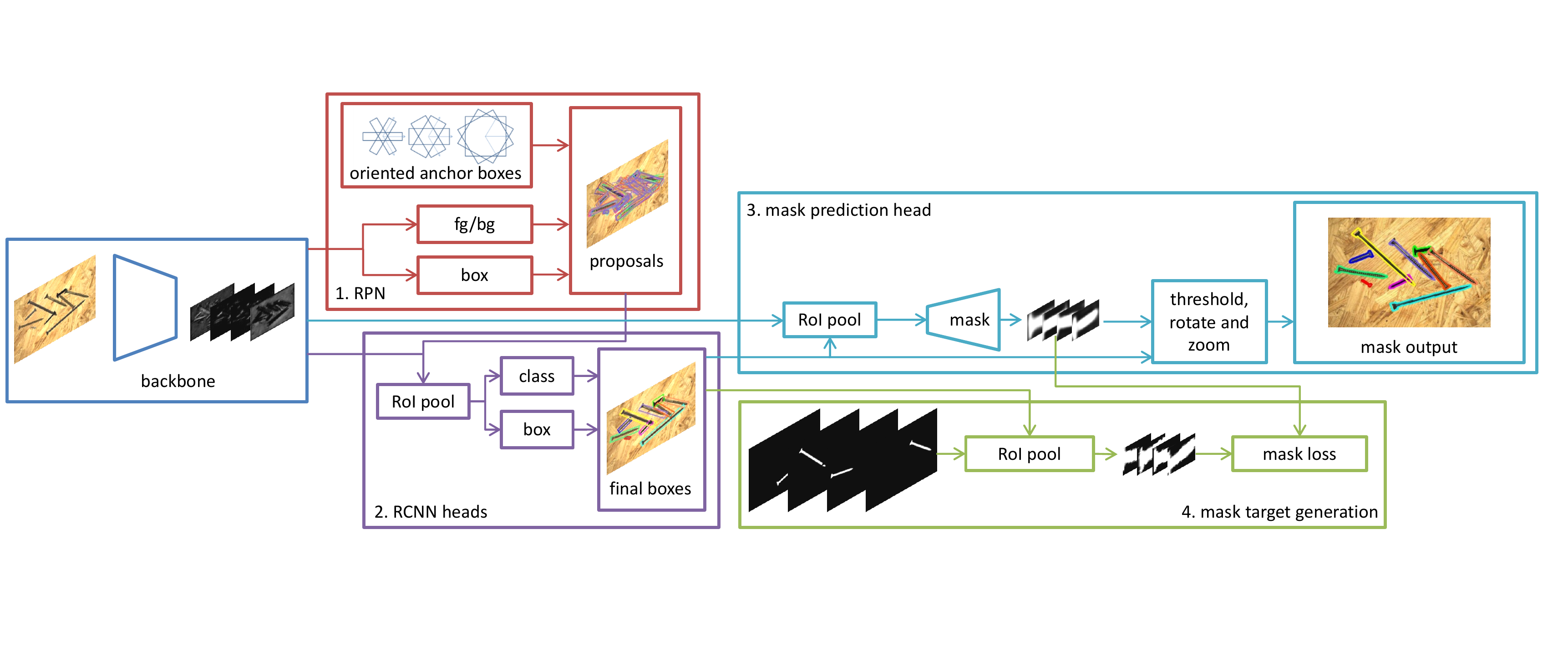}
  \caption{\small{\bf Exemplary \emph{OIS} Architecture.} (blue, center left): Input image and backbone for feature extraction. (red, top): Oriented RPN. (violet, bottom): Features are pooled with respect to oriented proposals to feed RCNN heads for final oriented box output. (cyan, middle): Features are pooled with respect to final boxes to feed the branch for mask prediction. Finally, mask probabilites are fit to the oriented box output and thresholded. (green, bottom-right): During training, mask targets are calculated by RoI pooling the GT masks.}
  \label{fig:architecture}
\end{figure*}
%%%%%%%%%%%%%%%%% Architecture Figure %%%%%%%%%%%%%%%%%%%

In the following, we present the key components to make the simple idea of \emph{OIS} work efficiently. Our approach is applicable to all \emph{AAIS} methods. An exemplary \emph{OIS} architecture based on Mask RCNN \cite{he_2017_mask} is depicted in \figref{fig:architecture}:
In a first step, the backbone is applied to the input image to extract features that are used for the three following stages: The first stage is the RPN, which predicts for each of a number of template \emph{anchor boxes} whether it is likely that an object with similar bounding box is present or not (fg/bg branch) and if so, how the anchor should be refined to better match the underlying object (box branch). The second stage are the RCNN heads, where the oriented box proposal outputs of the RPN are used to RoI pool the features for class prediction (class branch) and further box refinement (box branch). The third stage uses the final RCNN head oriented box output to again RoI pool the features to feed the mask prediction head that outputs a pixel-precise mask for each of the final boxes. All RoI pooling layers pool from the oriented grid aligned with the input boxes. Since the output feature maps are upright, usual convolutions can be used in the subsequent layers.

For an \emph{OIS} version of RetinaMask \cite{fu2019retinamask} the same strategy can be applied. The only difference is that the RCNN heads of the second stage are fused with the RPN stage that directly outputs class probabilities and the final boxes.

\paragraph{Box representation.} We use a five parameter representation for boxes $b = (r, c, l1, l2, \phi)$, where $(r,c)$ denote the subpixel-precise center point row and column coordinates, $(l1, l2)$ the semi-axes lengths of the oriented box, and $\phi$ the orientation pointing in the direction of the major axis. $\phi$ is given as an angle in radians between the positive horizontal axis and the $l1$-axis in mathematically positive sense. In contrast to the parametrization used in \cite{ma2018arbitrary}, we do not enforce that $l1 \geq l2$ such that sudden flips of the orientation for boxes with aspect ratio close to one are avoided.

%\paragraph{Backbone.} In \emph{OIS} we use a usual CNN classification network such as a ResNet \cite{He_2016_CVPR} or SqueezeNet \cite{iandola2016squeezenet} .

%%%%%%%%%%%%%%%%%% ArIoU Figure %%%%%%%%%%%%%%%%%%%
%\begin{figure}
%  \centering
%  \includegraphics[width=0.65\textwidth]{images/architecture/example_arIoU_and_box_target.pdf}
%  \caption{\small{\bf arIoU and box target example.} (left): (black) GT box, (red, blue) oriented anchors. Although, for the blue anchor the coordinates $l1, l2,$ and $\phi$ differ a lot, the exact IoU is higher than for the red anchor where only $\phi$ differs by $20^\circ$. This issue is solved by the use of the arIoU for anchor assignment. (right) Our box center targets ensure that the targets are consistent for anchors with different orientation but the same center-coordinates.}
%  \label{fig:arIoUandBoxTarget}
%\end{figure}
%%%%%%%%%%%%%%%%%% ArIoU Figure %%%%%%%%%%%%%%%%%%%

%%%%%%%%%%%%%%%%% Compare Mask Target Figure %%%%%%%%%%%%%%%%%%%
\begin{figure*}[t]
	\centering
	\begin{subfigure}[b]{\textwidth}
		\centering
		\includegraphics[width=0.1\textwidth]{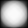}
		\includegraphics[width=0.1\textwidth]{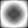}
		\hspace{0.2cm}
		\includegraphics[width=0.1\textwidth]{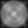}
		\includegraphics[width=0.1\textwidth]{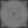}
		\hspace{0.2cm}
		\includegraphics[width=0.1\textwidth]{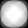}
		\includegraphics[width=0.1\textwidth]{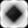}\\
		\centering
		\includegraphics[width=0.1\textwidth]{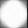}
		\includegraphics[width=0.1\textwidth]{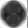}
		\hspace{0.2cm}
		\includegraphics[width=0.1\textwidth]{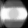}
		\includegraphics[width=0.1\textwidth]{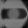}
		\hspace{0.2cm}
		\includegraphics[width=0.1\textwidth]{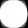}
		\includegraphics[width=0.1\textwidth]{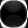}
		%\vspace{0.18cm}
		\caption{\small Mean and deviation of GT masks for \emph{D2S}, \emph{Screws} and \emph{Pill Bags} (\emph{left to right}), (\emph{top}) \emph{AAIS}, (\emph{bottom}) \emph{OIS}}
		\label{fig:mean_dev_masks}
	\end{subfigure}
	\begin{subfigure}[b]{0.24\textwidth}
		\centering
		\includegraphics[width=0.95\textwidth, clip, trim=0cm 0cm 15cm 0cm]{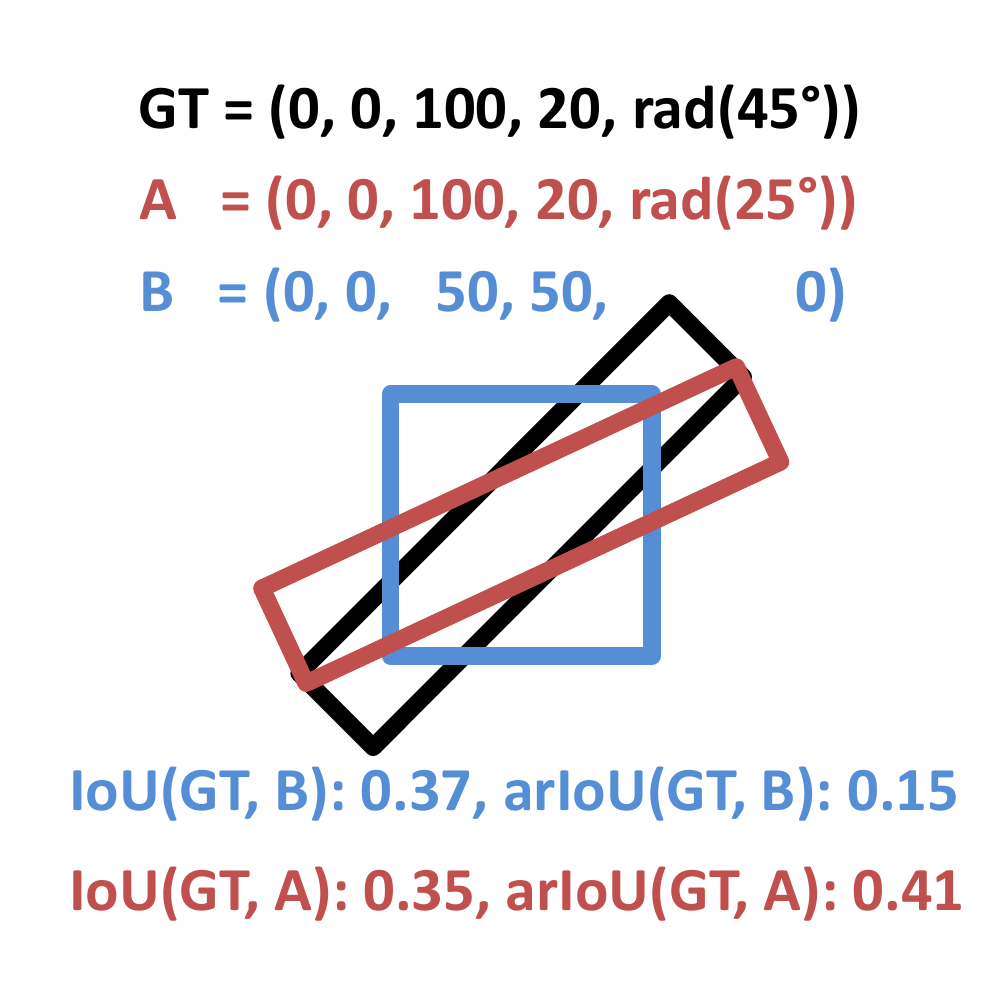}
		\caption{\small arIoU}
		\label{fig:arIoUandBoxTarget}
	\end{subfigure}
	\begin{subfigure}[b]{0.27\textwidth}
		\centering
		\includegraphics[width=0.17\textwidth]{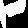}
		\includegraphics[width=0.17\textwidth]{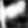}
		\includegraphics[width=0.17\textwidth]{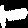}
		\includegraphics[width=0.17\textwidth]{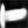} \\
		\includegraphics[width=0.17\textwidth]{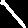}
		\includegraphics[width=0.17\textwidth]{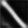}
		\includegraphics[width=0.17\textwidth]{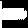}
		\includegraphics[width=0.17\textwidth]{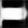}\\
		\includegraphics[width=0.17\textwidth]{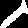}
		\includegraphics[width=0.17\textwidth]{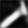}
		\includegraphics[width=0.17\textwidth]{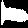}
		\includegraphics[width=0.17\textwidth]{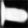}\\
		\includegraphics[width=0.17\textwidth]{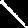}
		\includegraphics[width=0.17\textwidth]{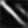}
		\includegraphics[width=0.17\textwidth]{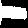}
		\includegraphics[width=0.17\textwidth]{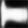} \\
		\vspace{0.2cm}
		\caption{\small Mask targets}
		\label{fig:MaskTargetsRect1Rect2}
	\end{subfigure}
	\begin{subfigure}[b]{0.47\textwidth}
		\centering
		\includegraphics[width=0.47\textwidth]{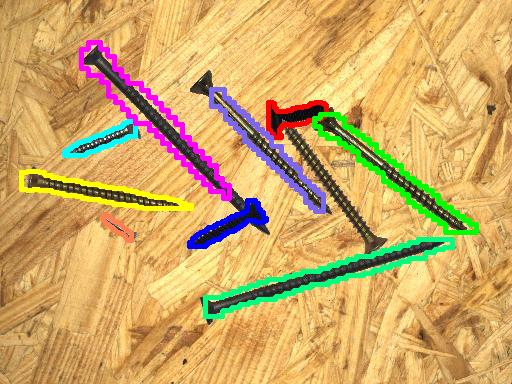}
		\includegraphics[width=0.47\textwidth]{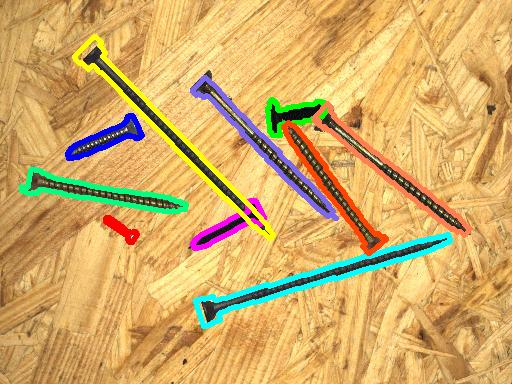}
		\vspace{0.18cm}
		\caption{\small Outputs: (\emph{left}) \emph{AAIS}, (\emph{right}) \emph{OIS}}
	\end{subfigure}
	
	\caption{\small{\bf Mean and deviation of GT masks, $\text{arIoU}$ \& comparison of mask targets and results.} (a) The mean and standard deviation of the first 3000 instance masks that are pooled to a size of $28 \times 28$ with respect to the GT boxes.
		(b) $\text{arIoU}$ example: (black) GT box, (red, blue) oriented anchors. Although, for the blue anchor the coordinates $l1, l2,$ and $\phi$ differ a lot, the exact IoU is higher than for the red anchor where only $\phi$ differs by $20^\circ$. This issue is solved by the use of the $\text{arIoU}$ for anchor assignment. (c) randomly chosen mask targets and mask probabilities, in the left two columns for \emph{AAIS}, in the right two columns for \emph{OIS}, respectively. (d) Final mask output for \emph{AAIS} and \emph{OIS} on a generated \emph{Screws} image (left and right)} %Note that the low resolution ($28\times28$) of mask targets and predictions is used much more efficiently for oriented boxes, where mask targets are more detailed. Moreover, the targets are much more consistent even for totally different instances. This simplifies both the training and the prediction.}
	\label{fig:MaskTarget}
\end{figure*}
%%%%%%%%%%%%%%%%% Compare Mask Target Figure %%%%%%%%%%%%%%%%%%%

\paragraph{RPN.} Because we want to predict oriented boxes, the region proposal network is fed with oriented anchors with different orientations, aspect ratios, and subscales. Ding \etal \cite{ding2019learning} argue that this leads to a massively higher number of anchors compared to axis-aligned boxes. However, we choose the orientation of the ground truth (GT) boxes such that it is aligned with the longer box side-length. Therefore, only aspect ratios smaller or equal to one are necessary, which almost halves the number of anchors. %Moreover, if one is not interested in the exact orientation of the box in the range $(-\pi, \pi]$, but only wants the angle to be exact modulo $\pi$ (range $(-\frac{\pi}{2}, \frac{\pi}{2}$), three orientations such as $(-\frac{\pi}{3}, 0, \frac{\pi}{3})$ are sufficient in most situations. 

As usual, the anchor target assignment is based on the IoU between anchors and GT: If an anchor has an IoU higher than $\textit{fgPosThresh}$ with a GT box it is assigned to the foreground and if the IoU is lower than $\textit{fgNegThresh}$ it is assigned to the background.
If the IoU is between $\textit{fgNegThresh}$ and $\textit{fgPosThresh}$ and there is another anchor that has a higher IoU with this GT, the anchor is ignored. However, if the IoU is higher than $\textit{fgNegThresh}$ and it is the highest IoU that was achieved by any anchor for this GT box, the anchor is assigned to foreground.

With a naive implementation we noticed that for \emph{OIS} the training is very unstable or converging to a bad local minimum due to the following reasons: On the one hand, the number of foreground anchors is very low or in some cases there is not a single anchor assigned to a particular GT instance. On the other hand, the exact oriented box IoU sometimes leads to very inefficient assignments (cf.\ \figref{fig:arIoUandBoxTarget}). 
To fix the latter, we replace the exact IoU by the angle-related IoU (arIoU \cite{liu2017learning}) for anchor assignment:
\begin{equation}
  \text{arIoU}(A, B) = \max(0, cos(\phi_A - \phi_B)) \cdot \text{IoU}(\hat{A}, B),
\end{equation}
where $\hat{A}$ is the box $(r_A, c_A, l1_A, l2_A, \phi_B)$.

Although the arIoU can fix the problem of wrong anchor assignment, the second problem of a low number of foreground instances remains because its values can still become small for anchors that have only slightly different parameters than the GT box. To fix this problem and increase the number of foreground anchors we decrease $\textit{fgNegThresh}$.
Moreover, we evaluate the influence of \emph{setting weak boxes to background (swb2bg)}: If \emph{swb2bg} is set to \emph{false}, anchors that achieve the highest arIoU with a GT box, but where the arIoU is below $\textit{fgNegThresh}$ are still assigned to the foreground.
By changing $\textit{fgNegThresh}$ and \emph{swb2bg} we assure that all GT instances contribute to the training, although none of the anchors fits perfectly. Hence, instead of increasing the number of anchors, these changes offer an efficient way to address all available GT box shapes. Effectively, this leads to a more stable training because enough foreground examples are present.

The relaxation of $\textit{fgNegThresh}$ and \emph{swb2bg} are only used in the RPN. For the RCNN heads and the mask head, we only want good box proposals to contribute to the training. This is in line with the findings of \cite{cai2018cascade}, where the assignment thresholds are increased in several iterative box refinements.

\paragraph{Oriented box targets.}

%The correct definition of training targets is a crucial ingredient for oriented box detection, as minor modifications can have a large impact on the training convergence.

We use the following box target calculations for the row and column coordinate deltas:
\begin{equation}
d_r = \frac{r^* - r}{\bar{l}}, ~ d_c = \frac{c^* - c}{\bar{l}},
\end{equation}
where $(r^*, c^*)$ are coordinates of the GT box, $(r, c)$ are coordinates, and $\bar{l} = (l1 + l2) / 2$ is the mean axis length of the box for which the deltas should be calculated. Using $\bar{l}$ for normalization there is no dependency on the orientation of the box. In previous work, such as \cite{ding2019learning,ma2018arbitrary}, $l2$ is used to normalize $d_r$ and $l1$ to normalize $d_c$ instead of the mean value $\bar{l}$. 
%This is similar to the original target normalization of \cite{girshick2014rich} for axis-aligned boxes with $\phi=0$.
However, if a box is oriented with $\phi \geq \frac{\pi}{2}$ the connotation of $l1$ and $l2$ with respect to $r$ and $c$ is flipped. This is avoided by the use of $\bar{l}$.

Moreover, in the calculation of the target delta for the orientation $d_\phi$, we avoid large regression targets by ensuring that $d_\phi$ is in the range $(-\frac{\pi}{2}, \frac{\pi}{2})$ due to the use of the arIoU during the assignment phase. %In practice, this works best if anchor orientations are chosen such that intermediate angles to the GT are not larger than $30^\circ$.

\paragraph{Oriented mask prediction.} 

For each predicted oriented box, features are pooled with an oriented RoI-pooling layer, such that the batch size of the mask branch equals the input batch size times the maximal number of predictions. The pooled feature maps with $m\times m = 14 \times 14$ spatial dimension are fed into a sequence of $n$ convolutions with ReLU activation followed by an upsampling transposed convolution. Finally, a sigmoid activation is applied to get the mask probability of each pixel within the warped box. Mask prediction is done class-agnostically since it has already been shown in \cite{he_2017_mask} that a class-specific mask prediction does not improve the results significantly but comes at a higher computational cost.

%Therefore, a higher mask resolution comes at the cost of much higher computation time and memory consumption and should be avoided.

For target generation, the GT masks are painted into a multi-channel binary image, where each channel contains the GT mask of one instance. Given the final box predictions,
the assignment to the GT instances is done in the same way as in the RPN based on the $\text{arIoU}$. In this separate assignment step, we increase both IoU thresholds such that only good final boxes contribute to the training of the mask branch. If a final box is not assigned to any GT instance, the corresponding weights in the mask prediction loss are set to zero such that the predicted mask is ignored.
We use a modified, oriented RoI-pooling on the multi-channel GT mask image that only pools within the channel indices of the assigned GT masks to obtain the mask probability targets for the predicted boxes (cf. bottom-right of \figref{fig:architecture}). A sigmoid cross-entropy loss is used to train the mask branch. 

As in \cite{he_2017_mask}, a final grid size of $28\times28$ pixels for mask targets and prediction is used. This low resolution limits the detailedness of the output masks. However, as the oriented boxes generally contain much less background than the axis-aligned boxes, the given resolution is used much more efficiently. This is visualized in \figref{fig:mean_dev_masks} and \figref{fig:MaskTargetsRect1Rect2}, where for \emph{OIS} the targets' and predictions' variance between different instances is significantly lower than for \emph{AAIS}. On the one hand, \emph{OIS} models can learn a stronger prior for the mask prediction since especially in the middle of the object it is very likely that the object is present. The capacity of the model can be focused on the boundary of the instances, where they differ the most. On the other hand, the mask prediction is done based on a higher resolution relative to the instance size. This gives the model more flexibility to capture fine details of the instance shape.

During inference, to obtain the final mask, the mask probabilities are thresholded by 0.5 and the obtained region is rotated, translated and zoomed according to the corresponding predicted box.

%As mentioned above, and in contrast to the original Mask RCNN implementation \cite{he_2017_mask}, we use the final oriented box predictions of the RCNN heads to pool features once again from the backbone and feed them into the mask branch.

%We incorporate the mask target generation directly into the model such that it can be done efficiently on the GPU during training. Therefore, we create a binary image where we paint the GT masks into zero-initialized channels one-by-one . The number of channels can be pre-calculated as the maximum number of instances per image. 
%To obtain the cropped mask targets, another oriented single-channel RoI pooling operation on the GT mask image is done. We adapt the RoI pooling operation  to pool from the assigned single channel of the GT mask image for each instance.

\paragraph{Classes without orientation.}

In many applications, not only objects with a clearly defined orientation are of interest. If we use the smallest oriented bounding box as GT for symmetric or round objects, the orientation is varying between different instances without a hint for the model why this is the case. This can lead to a destabilization of the training. Therefore, we propose to assign these classes to a group of \emph{classes without orientation}.

For these categories, we annotate the GT boxes as the smallest axis-aligned bounding boxes of the GT instance masks with $\phi=0$.

\paragraph{Implementation Details.}

We differentiate between two scenarios: The first scenario is that we are interested in the exact orientation of boxes, i.e., $\phi \in (-\pi, \pi]$. In the second scenario, $\phi$ is limited to the range $(-\frac{\pi}{2}, \frac{\pi}{2}]$, which is suitable if the actual orientation of the box is not important, but only a tight oriented bounding box of the instance mask is the goal. In this case, predictions with $\phi$ outside of the range can be corrected by subtracting or adding $\pi$.
To differentiate the two scenarios, we use a parameter \emph{IgnoreDirection} that is set to \emph{true} in the latter case.
Note that it is important that also the GT given in the dataset at hand is annotated in accordance to \emph{IgnoreDirection}.

Box-based instance segmentation models can be easily modified to be trainable only with mask GT available for a fraction of the training instances: Whenever the mask GT is not available for an instance, we set the loss weights of this instance within the mask branch to zero. Therefore, the prediction for this instance will be ignored during the loss computation.

%------------------------------------------------------------------------
\section{Experiments}

In the following subsections, we evaluate our \emph{OIS} model on three different datasets: namely the \emph{D2S} \cite{follmann2018d2s}, \emph{Screws} \cite{ulrich2019comparison} and the newly introduced \emph{Pill Bags} dataset. All datasets have high-quality annotations such that a potentially improved mask accuracy can be measured. %Additionally, they contain oriented classes as well as \emph{classes without orientation}, such that it can be shown that \emph{OIS} can handle both types at the same time.

To measure the potential benefit of \emph{OIS} compared to \emph{AAIS}, we compute the IoU of the GT mask with both the smallest oriented and axis-aligned bounding box for each instance. The mean box mask IoU shown in \tabref{tab:mBMIoU} contains the mean IoU values averaged over all classes. It shows how tighter the oriented boxes are on average. For all three datasets, the difference is significant and in the following we show that this potential can indeed be utilized.

\begin{table}[h]
	\centering
	%  \vspace{-0.8cm}
	\begin{tabular}{l | c | c}
		dataset & axis-aligned & oriented \\ \hline \hline
		\emph{D2S} & 62\% & 81\% \\
		\emph{Screws} & 46\% & 58\% \\
		\emph{Pill Bags} & 73\% & 84\% \\
	\end{tabular}
	\hfill
	\caption{\small{\bf Mean box mask IoU.} Mean IoU of smallest bounding box with mask computed for each instance of a class and averaged over all classes. %(\ref{eq:mBMIoU}). 
		For all three datasets, oriented bounding boxes approximate the mask a lot better than axis-aligned bounding boxes}
	\label{tab:mBMIoU}
\end{table}

\subsection{D2S}

The densely segmented supermarket dataset (\emph{D2S}) \cite{follmann2018d2s} contains 60 different categories of supermarket products lying on a turntable and captured in a top-down view with various orientations. There are elongated objects such as lying bottles, boxes, certain vegetables like carrot, or zucchini, but also round or highly deformable objects such as standing bottles, apples, or nets filled with oranges.

Because the instances are touching or overlapping in many cases, the dataset is very challenging. Moreover, difficult scenes are only contained in the validation and test splits, which requires data augmentation to enhance the training set.

In \emph{D2S}, each image is captured with three different lightings (normal, bright, and dark), but our focus is to show that \emph{OIS} improves the mask quality compared to the \emph{AAIS} baseline. Hence, we only use the normal lighting images. Further, we scale the images to a relatively low resolution of $512\times384$ to speed up the training and evaluation. %and to reduce the energy consumption of our experiments.

\paragraph{Data preparation and model settings.} To have a fair comparison, we use exactly the same training data for all models. As mentioned, for \emph{D2S} it is necessary to generate additional training data by utilizing the existing annotations of the training set. We follow the approach of \cite{follmann2018acquire} and crop random instances from the training set using their ground truth masks before pasting them onto random backgrounds that look similar to the ones of the validation and test set. %This way, the training set can be enlarged at almost no additional cost.
2000 additional augmented training images are generated with random backgrounds and randomly positioned and rotated instances. The dataset consists of 9000 images in total: 2000 augmented images plus 1960 original training images, 1200 validation images and 4340 test images. Examples for generated images are given in the supplementary material. For \emph{D2S}, we assign the following classes to \emph{classes without orientation}: \emph{apple\_golden\_delicious, apple\_granny\_smith, apple\_red\_boskoop, clementine, clementine\_single, orange\_single, oranges, lettuce, salad\_iceberg}. By their nature, these categories have no well-defined orientation. We set \emph{IgnoreDirection} to \emph{true} for \emph{D2S} experiments.

For the training, validation, and test splits, the box ground truth for \emph{OIS} is generated as follows: We use the smallest oriented bounding box of the instance mask by default, except for \emph{classes without orientation}, where the smallest axis-aligned bounding box is used. %Moreover, for instances, where the IoU of the smallest axis-aligned bounding box with the smallest oriented bounding box is higher than 0.95 and at the same time the roundness (cf.\ appendix) of the mask is higher than 0.95, we annotate the instance with the smallest axis-aligned bounding box.
%The same applies to the augmented images. However, because instances are frequently occluded, we use the orientation of the smallest oriented bounding box of the amodal mask \cite{li2016amodal} and fit the smallest bounding box with that orientation to the final, possibly occluded mask. This helps to obtain more consistent orientations and stabilizes the training.

For both RetinaMask (R) and Mask RCNN (M), we use the same network architecture for \emph{AAIS} and \emph{OIS}. For \emph{AAIS}, we replace oriented anchors with axis-aligned anchors and use the default axis-aligned RoI pooling operations. %Therefore, the intermediate box proposals generated by the RPN and the final box outputs generated by the RCNN heads are axis-aligned for \emph{AAIS}.
All other model hyperparameters are the same as for \emph{OIS}. In the RetinaMask experiments, we use two aspect ratios and three orientations for \emph{OIS} and six aspect ratios for \emph{AAIS} to get the same number of anchors. Interestingly the result could not be improved using more aspect ratios in the \emph{OIS} case.
%Due to the high complexity of the dataset, we use a ResNet-50 \cite{He_2016_CVPR} backbone pretrained on ImageNet \cite{deng_2009_imagenet}.
As backbone we use a ResNet-50 \cite{He_2016_CVPR} pretrained on ImageNet \cite{deng_2009_imagenet}. Note that we train all other weights of the RPN, the RCNN, and mask heads from scratch. We train the models for 50 (M) and 40 (R) epochs and evaluate after each epoch on the validation set (using box mAP). The results are always shown for the best model regarding the validation set mAP (early stopping) over 3 (M) and 1 (R) independent runs. All other training and model parameters can be found in the appendix.

\paragraph{Analysis.}

\begin{table}
  \centering
  \begin{tabular}{l | c | c | c | c}
    model        & AABB & OBB & mask & mask agn \\ \hline \hline
    \emph{AAIS} (M) & 51\%   & 9\%* & 45\%      & 49\% \\
    \emph{OIS} (M) &  52\%* & 48\%  & {\bf55\%} & 61\% \\
    \hline
    \emph{AAIS} (R) & {\bf 54\%} & 8\%* & 46\% & 55\% \\
    \emph{OIS} (R)  & {\bf 54\%}* & {\bf 50\%} & {\bf55\%} & {\bf65\%} \\
  \end{tabular}
  \caption{\small{\bf \emph{D2S} results.} mAP values on the test set. AABB: axis-aligned box, OBB: oriented box, mask: instance mask, mask agn: same as mask, but without evaluating the class prediction. \emph{OIS} clearly improves the mask mAP. (*) boxes are calculated from mask predictions}
  \label{tab:mAPD2S}
\end{table}

\begin{figure}[t]
	\centering
	 \setlength{\tabcolsep}{3pt}
	\begin{tabular}{c c c}
		input & \emph{AAIS} & \emph{OIS} \\
	\includegraphics[trim={0cm .337cm 0 0},clip,width=0.15\textwidth]{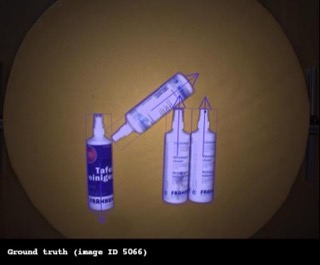}&
	\includegraphics[trim={0cm .337cm 0 0},clip,width=0.15\textwidth]{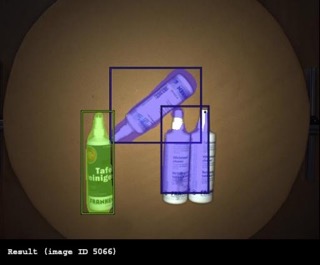}&
	\includegraphics[trim={0cm .337cm 0 0},clip,width=0.15\textwidth]{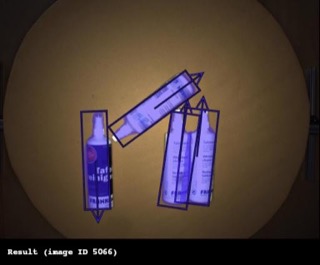}
    \\
	\includegraphics[trim={0cm .337cm 0 0},clip,width=0.15\textwidth]{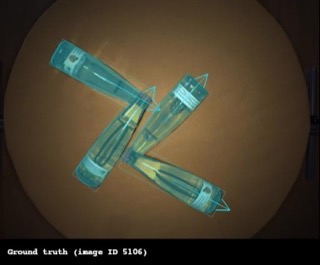}&
	\includegraphics[trim={0cm .337cm 0 0},clip,width=0.15\textwidth]{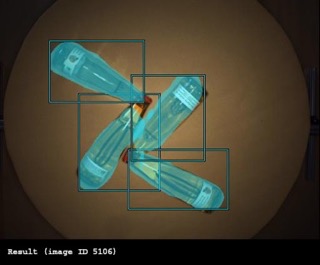}&
	\includegraphics[trim={0cm .337cm 0 0},clip,width=0.15\textwidth]{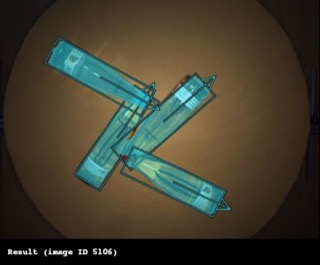}
	\\
	\includegraphics[trim={0cm .337cm 0 0},clip,width=0.15\textwidth]{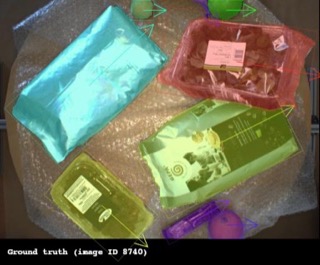}&
	\includegraphics[trim={0cm .337cm 0 0},clip,width=0.15\textwidth]{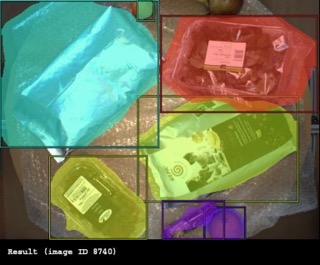}&
	\includegraphics[trim={0cm .337cm 0 0},clip,width=0.15\textwidth]{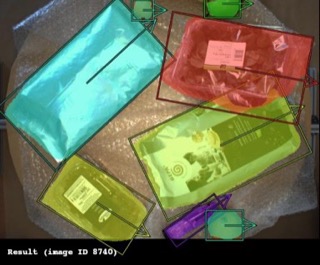}
	\\
	\includegraphics[trim={0cm .337cm 0 0},clip,width=0.15\textwidth]{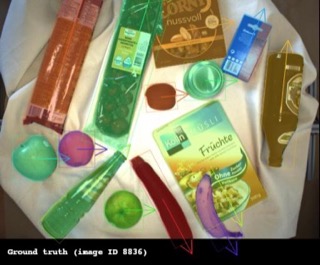}&
	\includegraphics[trim={0cm .337cm 0 0},clip,width=0.15\textwidth]{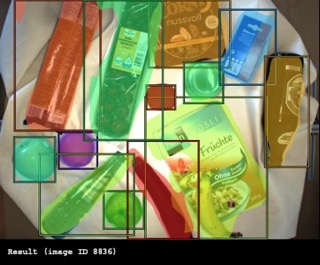}&
	\includegraphics[trim={0cm .337cm 0 0},clip,width=0.15\textwidth]{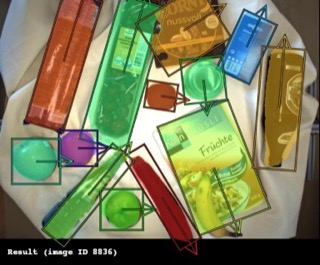}
	\\
	\includegraphics[trim={0cm .337cm 0 0},clip,width=0.15\textwidth]{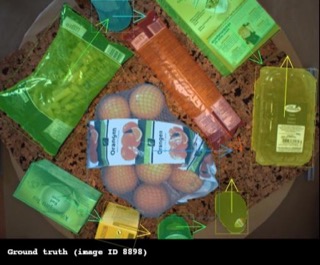}&
	\includegraphics[trim={0cm .337cm 0 0},clip,width=0.15\textwidth]{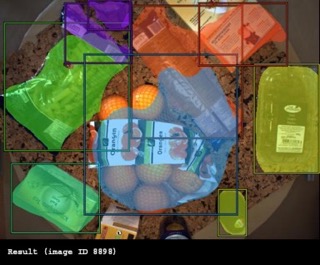}&
	\includegraphics[trim={0cm .337cm 0 0},clip,width=0.15\textwidth]{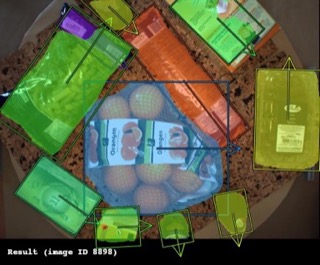}
	\end{tabular}
	\caption{\small \textbf{\emph{D2S} results.} \emph{(from left to right)} input image with ground truth masks and generated ground truth oriented boxes, results of \emph{AAIS}, results of \emph{OIS}. \emph{OIS} clearly improves the mask accuracy, especially for close-packed, elongated, and diagonally oriented objects. More results can be found in the supplementary material (best viewed digitally with zoom)}
	\label{fig:D2SResImages}
\end{figure}

\figref{fig:D2SResImages} shows qualitative results on \emph{D2S} test images. In comparison to \emph{AAIS}, the \emph{OIS} mask predictions are much more accurate, especially for diagonally oriented objects. Due to the less precise mask targets in \emph{AAIS}, also the mask probabilities become inexact. In combination with the upscaling to the final box size and thresholding, the mask output is often not exactly in correspondence with the instance boundaries. In case of overlapping or touching objects, \emph{OIS} clearly shows better results as in those cases the box overlap is lower. In many cases for \emph{AAIS}, the mask is extended into the background or onto neighboring objects.

\begin{figure}[t]
  \centering
  % TIKZ %%%%%%%%%%%%%%%%%%%
  \begin{tikzpicture}
  \begin{axis}[
  scale only axis=true,
  width=6.5cm, height=3.0cm,
  x label style={at={(axis description cs:0.5,-0.15)},anchor=north},
  y label style={at={(axis description cs:-0.15,.5)},anchor=south},
  xlabel={IoU-threshold},
  ylabel={mAP},
  xmin=0.45, xmax=1.0,
  ymin=0, ymax=0.85,
  xtick={0.5,0.6,0.7,0.8, 0.9},
  ytick={0,.20,.40,.60,.80},
  legend style={draw=none},
  legend pos = outer north east,
  legend columns=1,
  ymajorgrids=true,
  grid style=dashed,
  axis lines=left,
  every axis plot/.append style={thick}
  ]
  
  \addplot [color=colorRect2] table {plots/mAPs_d2s_sr2_cls_specific_rect2_new.dat};
  %\addplot [color=colorRect2Agnostic] table {plots/mAPs_d2s_sr2_cls_agnostic_rect2.dat};
  \addplot [color=colorRect1] table {plots/mAPs_d2s_cls_specific_rect1_new.dat};
  %\addplot [color=colorRect1Agnostic] table {plots/mAPs_d2s_cls_agnostic_rect1.dat};
  %\legend{{\tt OIS}, {\tt OIS agn}, {\tt AAIS}, {\tt AAIS agn}}
  \legend{\emph{OIS}, \emph{AAIS}}
  \end{axis}
  \end{tikzpicture}
  % TIKZ %%%%%%%%%%%%%%%%%%%
  \hfill
  \caption{\small {\bf Mask RCNN on \emph{D2S}.} Mask mAP-values for IoU-thresholds from 0.5 to 0.95. Consistently and especially for high IoU-thresholds \emph{OIS} improves the mAP significantly}
  \label{fig:mAPD2S}
\end{figure}
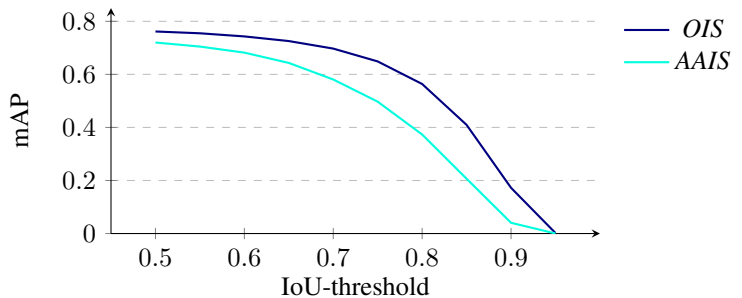

Generally, if the box is predicted correctly, the mask is very precise for \emph{OIS}. This is a clear improvement to \emph{AAIS} consistently across RetinaMask and Mask RCNN, which also reflects in mAP-values shown in \tabref{tab:mAPD2S}. \figref{fig:mAPD2S} shows that especially for high IoU-thresholds, the mAP can be significantly improved by \emph{OIS}: E.g., from 50\% to 65\% @IoU0.75 and from 21\% to 41\% @IoU0.85.

Since \emph{AAIS} predicts axis-aligned boxes, we can only evaluate the box mAP for the axis-aligned ground truth boxes. Interestingly, if we use the axis-aligned smallest bounding boxes of the instance masks predicted by \emph{OIS} (M), the mAP of 55\% outperforms the box mAP of \emph{AAIS} (M). In contrast, calculating the oriented box from the \emph{AAIS} mask does not work at all. Moreover, in contrast to \emph{AAIS}, for \emph{OIS} the mask mAP is even higher than the oriented box mAP, which shows that the mask prediction is very precise.
Since our focus is to improve mask accuracy, we also show results where the predicted class is not taken into account (class-agnostic evaluation). Also in this case, the mask mAP can be improved substantially.

%%%%%%%%%%%%%%%%% FIGURE IoU Mask rect1 vs. rec2 %%%%%%%%%%%%%%%%%%%
%%%%%%%%%%%%%%%%% FIGURE %%%%%%%%%%%%%%%%%%%

\begin{table}
    \centering
    \begin{tabular}[h]{c | c | c | c | c}
   $\neg$\emph{swb2bg} & \emph{fgNegThresh} & \emph{cwo} & OBB & mask \\ \hline
   & &             & 31.56\% $\pm$ 0.004 & 40.13\% $\pm$ 0.002 \\
   \checkmark  & &    & 32.58\% $\pm$ 0.004 & 40.85\% $\pm$ 0.001 \\
  % & \checkmark & & 45.1\% & 53.6\% \\
  \checkmark& \checkmark &   & 44.84\% $\pm$ 0.001 & 53.39\% $\pm$ 0.001 \\
  % & \checkmark  & \checkmark & 47.33\% $\pm$ 0.001 & 53.47\% $\pm$ 0.009 \\
  \checkmark & \checkmark &   \checkmark & {\bf 47.01}\% $\pm$ 0.012 & {\bf 54.06}\% $\pm$ 0.008  
  \end{tabular}
  \caption{\small{\bf Ablation study for \emph{D2S}}. mean Mask mAP values on test set over three independent runs and their standard deviation. Increasing the number of foreground examples in training by decreasing \emph{fgNegThresh} and disabling \emph{swb2bg} improves the results. Setting \emph{cwo} additionally improves the mask mAP}
  \label{tab:ablationstudyD2S}
  %\vspace{-.5cm}

\end{table}

\paragraph{Ablation study.}

We also perform an ablation study for \emph{OIS} to empirically confirm our hyperparameter settings. Results are shown in \tabref{tab:ablationstudyD2S}. Namely, we evaluate the influence of reducing \emph{fgNegThresh} from 0.4 to 0.3 and switching off \emph{setWeakBoxesToBg (swb2bg)}, as well as using \emph{classes without orientation (cwo)}. The results indicate that \emph{swb2bg} and \emph{cwo} only slightly improve the mask mAP and the most benefit comes from reducing \emph{fgNegThresh}. Using \emph{cwo} mainly improves the box mAP because this assures that for symmetric or round objects the orientation of the bounding box is well defined. Overall, the mask mAP can be drastically improved from 40\% to 54\%.

\subsection{Screws}

\begin{figure}[t]
        \centering
        \setlength{\tabcolsep}{3pt}
        \begin{tabular}{c c c}
        	input & \emph{AAIS} & \emph{OIS} \\
        \includegraphics[trim={0cm .337cm 0 0},clip,width=0.15\textwidth]{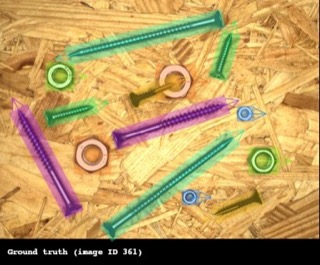}&
        \includegraphics[trim={0cm .337cm 0 0},clip,width=0.15\textwidth]{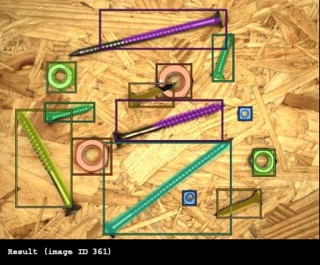}&
        \includegraphics[trim={0cm .337cm 0 0},clip,width=0.15\textwidth]{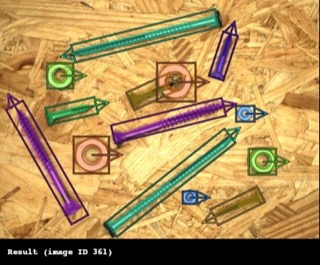}
        \\
        \includegraphics[trim={0cm .337cm 0 0},clip,width=0.15\textwidth]{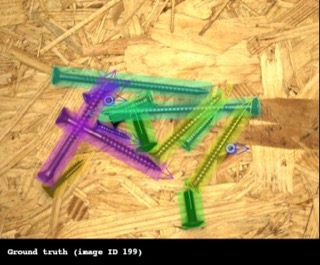}&
        \includegraphics[trim={0cm .337cm 0 0},clip,width=0.15\textwidth]{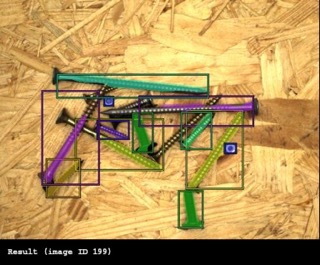}&
        \includegraphics[trim={0cm .337cm 0 0},clip,width=0.15\textwidth]{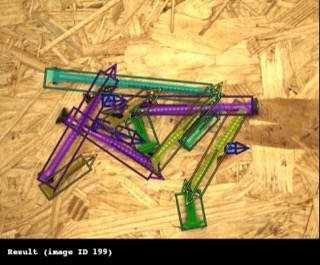}
        \\
        \includegraphics[trim={0cm .337cm 0 0},clip,width=0.15\textwidth]{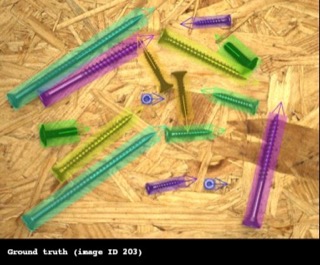}&
        \includegraphics[trim={0cm .337cm 0 0},clip,width=0.15\textwidth]{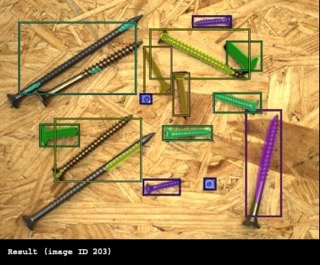}&
        \includegraphics[trim={0cm .337cm 0 0},clip,width=0.15\textwidth]{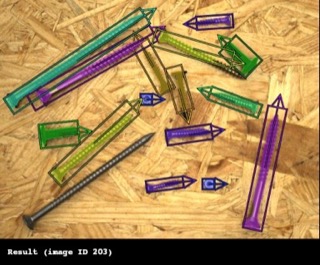}
        \\
        \includegraphics[trim={0cm .337cm 0 0},clip,width=0.15\textwidth]{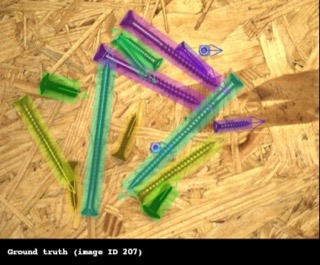}&
        \includegraphics[trim={0cm .337cm 0 0},clip,width=0.15\textwidth]{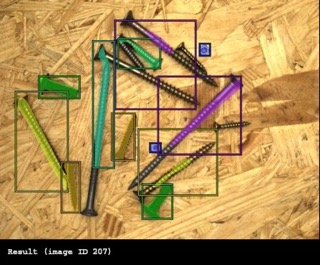}&
        \includegraphics[trim={0cm .337cm 0 0},clip,width=0.15\textwidth]{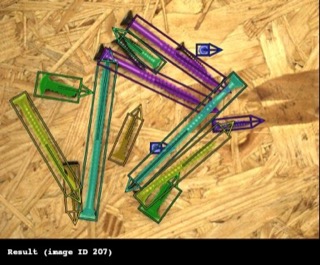}
        \\
        \includegraphics[trim={0cm .337cm 0 0},clip,width=0.15\textwidth]{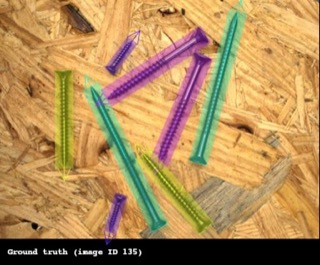}&
        \includegraphics[trim={0cm .337cm 0 0},clip,width=0.15\textwidth]{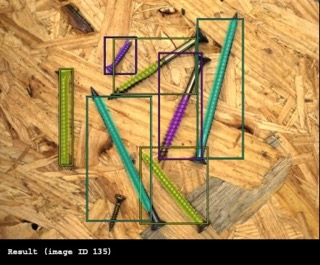}&
        \includegraphics[trim={0cm .337cm 0 0},clip,width=0.15\textwidth]{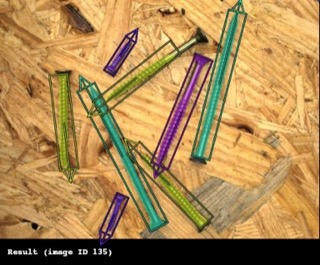}
        \end{tabular}

  \caption{\textbf{\emph{Screws} results.} \emph{(from left to right)} real input with ground truth boxes, results of \emph{AAIS}, results of \emph{OIS}. Models were trained only on generated data. \emph{OIS} clearly improves the mask accuracy, especially for close-packed, elongated, and diagonally oriented objects. More results can be found in the supplementary material (best viewed digitally with zoom)}
          \label{fig:ScrewsResImages}
\end{figure}

The \emph{Screws} \cite{ulrich2019comparison} dataset contains 9 different types of screws and 4 different types of nuts on a wooden background. The categories differ in size, length, and color, but some are only distinguishable by a different thread. Typical example images are shown in \figref{fig:ScrewsResImages}.

The dataset consists of 384 images, of which 70\% belong to the train set, 15\% to the validation set, and 15\% to the test set. Ground truth annotations are given as oriented boxes. As the nut-classes are symmetric and the orientation is not well-defined, these four classes are assigned to \emph{classes without orientation}. In this dataset for screw classes, the exact orientation pointing from the screws head to its tail is of interest. Therefore, \emph{IgnoreDirection} is set to \emph{false} such that box orientations are in the range $(-\pi, \pi]$.

\paragraph{Generated training data.} Because the pixel-precise annotation of instance masks is tedious and time-consuming, we follow the approach of \cite{ulrich2019comparison} to generate artificial training images. Therefore, each category is captured on a homogeneous white background where a relatively precise instance mask can be obtained by thresholding. In this weakly-supervised setting, we use a single template image per category and generate 800 images (350 train, 150 validation, 300 test) by cropping and pasting random instances onto empty wooden backgrounds similar to those of the original dataset. A generated example image is shown in \figref{fig:MaskTarget}.

To train and evaluate \emph{AAIS}, we obtain the box ground truth annotations as the smallest axis-aligned bounding box of instance masks. 
The oriented box ground truth for categories with orientation is generated as follows: The orientation of the box is calculated based on the second moments of the instance mask (cf. appendix). To get the orientation pointing from the screws head to its tail, the orientation is corrected by adding or subtracting $\pi$ if the orientation from the center of gravity to the most distant point on the mask boundary is pointing into the opposite direction.

\paragraph{Model settings.} We train all models only on the generated training set. Because \emph{Screws} has less intra-class variations and less variations due to changes in perspective, the dataset is less complex than \emph{D2S}. Therefore, in \emph{Screws} experiments we use a SqueezeNet \cite{iandola2016squeezenet} backbone, which speeds up the training process. This also shows that the proposed \emph{OIS} model is superior independent of the backbone or other hyperparameters shown in the appendix.

\paragraph{Evaluation on generated data.} Quantitative results are shown in \tabref{tab:mAPScrews}: First, we evaluate \emph{AAIS} and \emph{OIS} on the generated test set, where the box mAP is very similar for \emph{AAIS} and \emph{OIS} (55.2\% vs. 58.7\%). %The reason why \emph{OIS} has a higher box mAP is that for oriented detection generally the bounding boxes of overlapping instances do not overlap as much as for \emph{AAIS}. Thus, on the one hand, the pooled features contain less influences from other instances. On the other hand, fewer of the correct predictions are filtered out due to NMS.
Also here, the clear improvement of mask mAP (41.5\% \emph{AAIS} to 53.4\% \emph{OIS}) shows that the masks can be predicted much more precise for \emph{OIS}.

\begin{figure}
  \centering
%  \vspace{-0.4cm}
  \begin{minipage}{0.47\textwidth}
  \centering
  \begin{tabular}{l | c | c | c}
    model & AABB & OBB & mask \\ \hline \hline
    \multicolumn{4}{c}{\emph{generated data}} \\ \hline
    \emph{AAIS} & 55.2\% & -       & 41.5\% \\
    \emph{OIS}  & -      &  58.7\% & {\bf 53.4\% } \\ \hline
    \multicolumn{4}{c}{\emph{real data}} \\ \hline
    \emph{AAIS bfm} & - & 36.1\% & - \\
    \emph{OIS bfm}  & - & 45.6\% & - \\
    \emph{OIS}                & - & 51.4\% & - \\
    \emph{OIS ofm}   & - & {\bf 52.9\%} & - \\
  \end{tabular}
  \captionsetup{labelformat=table}
  \caption{\small{\bf \emph{Screws} Mask RCNN results.} mAP values on the test sets. AABB: axis-aligned box result, OBB: oriented box result, mask: instance mask result. \emph{OIS} clearly improves the mask mAP and the generated masks are a lot more consistent with the instances}
  \label{tab:mAPScrews}
\end{minipage}
\hspace{0.2cm}
\begin{minipage}{0.47\textwidth}
    \centering
%    \begin{figure}[t]
        \centering
        \begin{tikzpicture}
        \begin{axis}[
        scale only axis=true,
        width=4.0cm, height=2.5cm,
        ymin=0.7, ymax=0.85, xmin=5.0, xmax=105.0,
        ytick={0.7,0.75,0.8,0.85}, ytick align=outside, ytick pos=left,
        xtick={10,30,50,70,90,100}, xtick align=outside, xtick pos=left,
        xlabel={train boxes with mask GT [\%]},
        ylabel={mAP on test set},
        axis lines=left,
        ymajorgrids=true,
        grid style=dashed,
        legend pos=south east,
        legend style={draw=none},
        every axis plot/.append style={thick}]
        \addplot+[
        colorRect2, mark options={colorRect2, scale=0.75},
        error bars/.cd, 
        y fixed,
        y dir=both, 
        y explicit
        ] table [x=x, y=y,y error=error, col sep=comma] {plots/pill_bags_rect2_per_fraction.dat};
        \addlegendentry{\emph{OIS}}
        \addplot+[
        colorRect1, mark options={colorRect1, scale=0.75},
        error bars/.cd, 
        y fixed,
        y dir=both, 
        y explicit
        ] table [x=x, y=y,y error=error, col sep=comma] {plots/pill_bags_rect1_per_fraction.dat};
        \addlegendentry{\emph{AAIS}}
        
        \end{axis}
        \end{tikzpicture}
        \caption{\small{\bf \emph{Pill bags} partial supervision.} For \emph{OIS}, if only 10\% of the boxes contain a mask ground truth, the mAP is on the same level as with all mask annotations. \emph{AAIS} needs at least 50\% of the boxes labeled with instance masks} 
        \label{fig:results_pill_bags}
%    \end{figure}
\end{minipage}
%\vspace{-0.8cm}
\end{figure} 

%%%%%%%%%%%%%%%%% Screws Results Figure %%%%%%%%%%%%%%%%%%%
%%%%%%%%%%%%%%%%% Screws Results Figure %%%%%%%%%%%%%%%%%%%

\paragraph{Evaluation on real data.}

Because the original \emph{Screws} dataset only contains oriented box annotations, we predict instance masks and use them to calculate oriented boxes in the same way as was done for the ground truth of the generated dataset (box from mask \emph{bfm}). By this, we measure how well the predicted masks are aligned with the instance. For \emph{OIS}, we can also evaluate the predicted oriented boxes. The result of \emph{OIS ofm} is using the predicted box from \emph{OIS}, where $\phi$ is changed to the orientation calculated from the predicted mask.

\tabref{tab:mAPScrews} shows that for \emph{AAIS} the predicted masks do not coincide with the instances, as the box mAP of 36.1\% is rather low. For \emph{OIS}, this result is increased by a large margin to 45.6\%. The result is still slightly below the box result directly predicted from \emph{OIS}. However, if predicted masks are used to refine the orientation as in \emph{OIS ofm} the box results can be further improved by 1.5\%.
Qualitative results are shown in \figref{fig:ScrewsResImages}.

\subsection{\emph{Pill Bags}}

The \emph{Pill Bags} dataset contains 398 images of 10 different pill categories captured within a plastic bag. Overall the dataset contains 4048 instances, whereof 2826, 622, and 600 belong to the 278, 61, and 59 images of the training, validation and test sets, respectively. As in a typical industrial application, all images have more or less the same appearance. Moreover, the pills have not many deformations and their shape is close to that of an ellipse. Therefore, it is not surprising that the instance segmentation task can already be solved quite well with \emph{AAIS}. However, in our experiments on \emph{Pill Bags}, we analyze how many of the instance mask labels are necessary to obtain a good result. Therefore, we delete all but a fraction of the annotated GT instance masks and whenever the mask is not available during training, we set the corresponding mask loss weight of this instance to zero. This allows to train only with a fraction of the expensive instance mask labels, which are tedious to annotate. %This is similar to the \emph{class-agnostic} setting of \cite{hu2018learning}.

\begin{figure}[ht]
    \centering
    \setlength{\tabcolsep}{3pt}
    \begin{tabular}{c c c}
    	input & \emph{AAIS} & \emph{OIS} \\
    \includegraphics[trim={0cm .337cm 0 0},clip,width=0.15\textwidth]{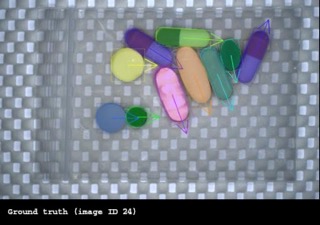}&
     \includegraphics[trim={0cm .337cm 0 0},clip,width=0.15\textwidth]{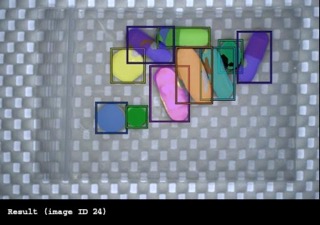}&
     \includegraphics[trim={0cm .337cm 0 0},clip,width=0.15\textwidth]{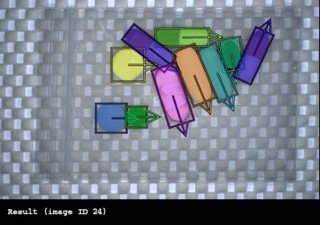} \\
     %
     %\hspace{0.1cm}
     \includegraphics[trim={0cm .337cm 0 0},clip,width=0.15\textwidth]{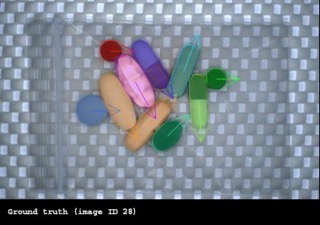}&
     \includegraphics[trim={0cm .337cm 0 0},clip,width=0.15\textwidth]{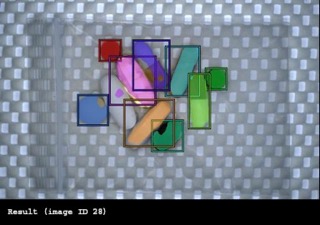}&
     \includegraphics[trim={0cm .337cm 0 0},clip,width=0.15\textwidth]{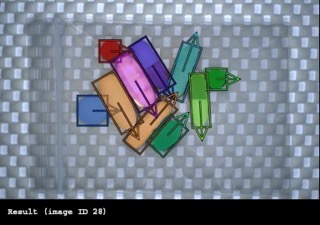} \\
     %
     %\includegraphics[trim={0cm .337cm 0 0},clip,width=0.15\textwidth]{images/results/pill_bags/rect2_10_gt_0008.jpg}
     %\includegraphics[trim={0cm .337cm 0 0},clip,width=0.15\textwidth]{images/results/pill_bags/rect1_10_result_0008.jpg}
     %\includegraphics[trim={0cm .337cm 0 0},clip,width=0.15\textwidth]{images/results/pill_bags/rect2_10_result_0008.jpg} \\
     %
     %\hspace{0.1cm}
     \includegraphics[trim={0cm .337cm 0 0},clip,width=0.15\textwidth]{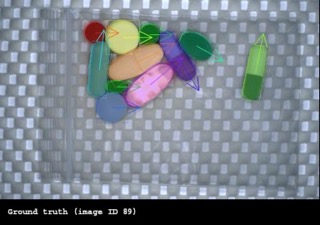}&
     \includegraphics[trim={0cm .337cm 0 0},clip,width=0.15\textwidth]{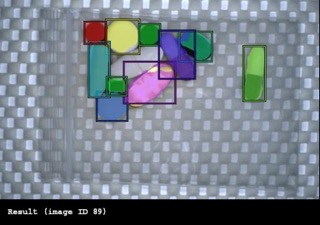}&
    \includegraphics[trim={0cm .337cm 0 0},clip,width=0.15\textwidth]{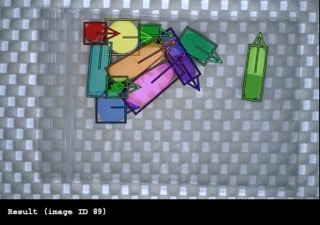}
     \end{tabular}
    \caption{\textbf{\emph{Pill Bags} results.} \emph{(from left to right)} input image with GT masks and GT oriented boxes, results of \emph{AAIS} (R), results of \emph{OIS} (R). Both models have been trained with only 10\% of GT instance masks available.}
    \label{fig:PillBagsResImages}
\end{figure}

\figref{fig:results_pill_bags} shows that for \emph{OIS} we can already successfully train instance segmentation if only 10\% of the masks are annotated in the training set - without a significant reduction in mask mAP on the test set. In comparison, the \emph{AAIS} method significantly benefits from more annotations. Even using all annotations the \emph{AAIS} resulting mAP is lower than the 10\% counterpart of \emph{OIS}. This can be explained by the higher variance of the mask within the axis-aligned bounding boxes. Interestingly, for both methods the best results can be obtained when only a fraction of the instances are annotated. Some qualitative examples are shown in \figref{fig:PillBagsResImages}.

%\vspace{-1cm}

%------------------------------------------------------------------------
\section{Conclusion}

In this work, we have presented an instance segmentation model that predicts instance masks based on oriented box predictions. The use of oriented boxes leads to a more precise approximation of the instance masks and results in very accurate mask predictions.
We have shown that the overall mask mAP on \emph{D2S} and \emph{Screws} can be improved significantly. The predicted instance masks are very accurate, such that they can be used to improve the axis-aligned box detection on \emph{D2S} and the oriented box detection on \emph{Screws}.
Moreover, oriented instance segmentation on \emph{Pill bags} requires only 10\% of instance mask annotations.

{\small
\bibliographystyle{plain}
\bibliography{ois}
}

\appendix

We provide the following supplementary material:
\begin{itemize}
  \item Per instance evaluations,
  \item more detailed qualitative results,
  \item calculation of orientation of a region,
  \item detailed model configurations,
  \item detailed solver settings.
  %\item videos containing results for test images with a per instance comparison.
\end{itemize}

\section{Per Instance Evaluations}

In our paper, we show that for \emph{D2S} and \emph{Screws} (generated), we get large relative mask mAP improvements. Because the mAP depends not only on the accuracy of the mask, but also on the class, in this section we evaluate the mean per instance IoU of predicted masks. Therefore, for each ground truth instance, we calculate the maximum IoU with respect to all predicted instances, independent of the predicted class. We further compute the mean IoU, once over all ground truth instances (all), and once only over instances where the achieved maximum IoU is larger than zero ($>$0). We also compute the number of instances that result as false negatives because the maximum IoU is below 0.75 (NumFNIoU (@0.75)).

Further, we check if the predicted class of the mask that achieves the maximum IoU for a ground truth instance, is correct. We sum these correctly predicted classes over the whole test dataset in order to see how much the mAP improvement depends on the class prediction.

The results are summarized in \tabref{tab:MeanIoU}. The mean IoU is improved significantly for \emph{OIS} compared to \emph{AAIS} (both all and $>$0). This results in a lot less false negatives at IoU threshold 0.75. For \emph{Screws}, the number of predictions with correct class is slightly worse for \emph{OIS} than for \emph{AAIS}. This emphasizes that for \emph{Screws} the mAP improvement is coming from more accurate mask predictions. For \emph{D2S}, the number of predictions with correct class is slightly increased by \emph{OIS}, but still on the same level (81.3\% \emph{OIS} vs. 79.8\% \emph{AAIS}). Hence, also here the mAP improvement is mainly based on more accurate masks.

\begin{table}[b]
	\centering
	\small
	\begin{tabular}{l | c  c | c  c} \hline
		& \multicolumn{2}{c|}{\emph{Screws} (generated)} & \multicolumn{2}{c}{\emph{D2S}} \\ 
		& \emph{AAIS} & \emph{OIS} & \emph{AAIS} & \emph{OIS} \\\hline \hline
		Num GT     & \multicolumn{2}{c|}{3121} & \multicolumn{2}{c}{16630} \\ \hline
		Mean IoU (all) & 68.9\% & 73.8\% & 73.0\% & 78.5\% \\
		Mean IoU ($>$0)  & 70.2\% & 76.0\% & 74.3\% & 79.8\% \\
		NumFNIoU (@0.75) & 1844 & 1233 & 5482 & 3547 \\
		NumClassOk & 2990 & 2998 & 13287 & 13522 \\ \hline
	\end{tabular}
	\caption{{\bf Mean IoU evaluations.} NumGT: Number of ground truth instances in test dataset. Mean IoU (all): mean over maximum IoU values achieved for all ground truth instances. Mean IoU ($>$0): mean over maximum IoU values achieved for all ground truth instances where maximum IoU $>$ 0. NumFNIoU (@0.75): number of ground truth instances, where maximum IoU $<$ 0.75. NumClassOk: Number of predictions that achieved the maximum IoU value and where the predicted class was correct. See text for interpretation}
	\label{tab:MeanIoU}
\end{table}

\section{More Detailed Qualitative Results}

\begin{figure*}
	\begin{center}
		\setlength\tabcolsep{2pt}
		\begin{tabular}{c c c c}
			input & GT box and mask & \emph{AAIS} & \emph{OIS} \\
			% FIRST row
			\includegraphics[width=0.23\textwidth]{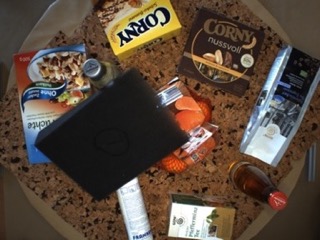}&
			\includegraphics[width=0.23\textwidth]{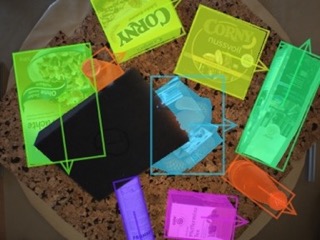}&
			\includegraphics[width=0.23\textwidth]{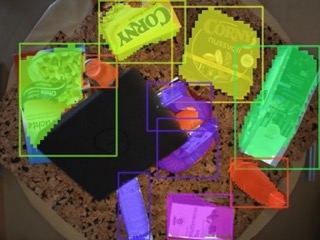}&
			\includegraphics[width=0.23\textwidth]{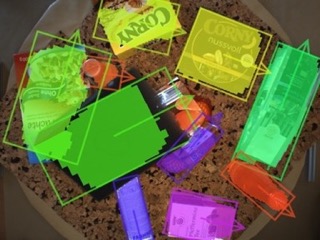}\\
			\includegraphics[width=0.23\textwidth]{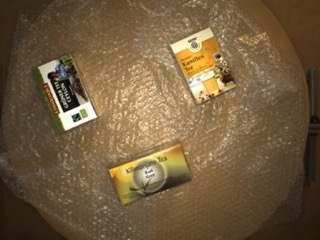}&
			\includegraphics[width=0.23\textwidth]{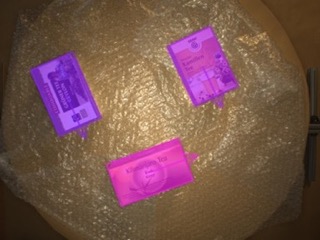}&
			\includegraphics[width=0.23\textwidth]{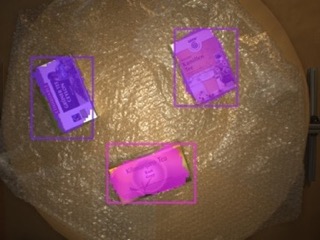}&
			\includegraphics[width=0.23\textwidth]{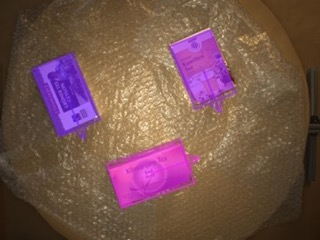}\\
			\includegraphics[width=0.23\textwidth]{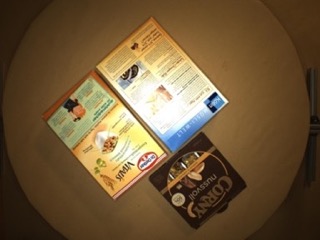}&
			\includegraphics[width=0.23\textwidth]{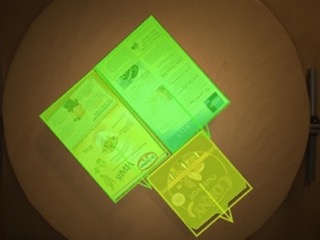}&
			\includegraphics[width=0.23\textwidth]{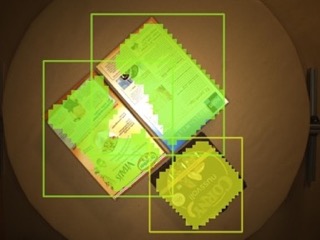}&
			\includegraphics[width=0.23\textwidth]{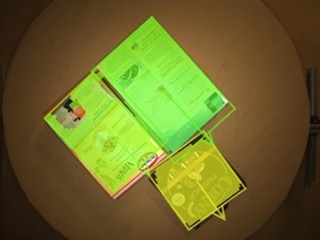}\\
			\includegraphics[width=0.23\textwidth]{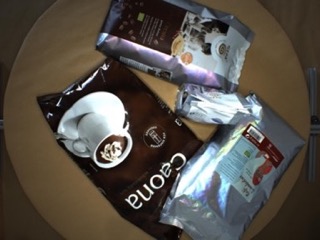}&
			\includegraphics[width=0.23\textwidth]{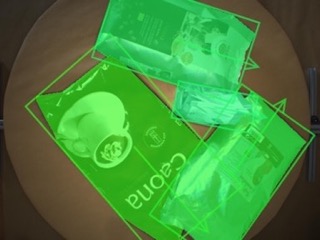}&
			\includegraphics[width=0.23\textwidth]{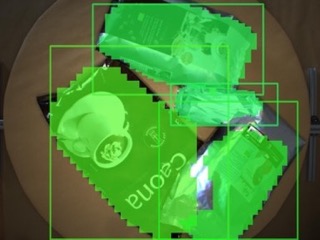}&
			\includegraphics[width=0.23\textwidth]{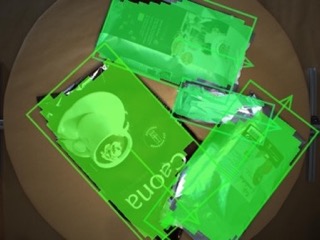}\\
			\includegraphics[width=0.23\textwidth]{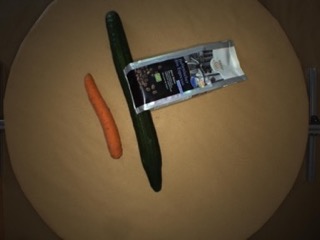}&
			\includegraphics[width=0.23\textwidth]{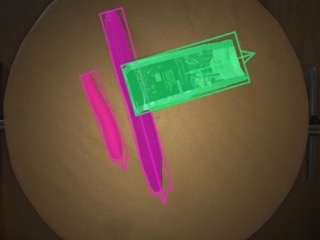}&
			\includegraphics[width=0.23\textwidth]{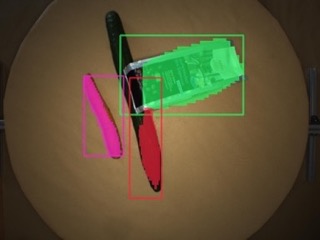}&
			\includegraphics[width=0.23\textwidth]{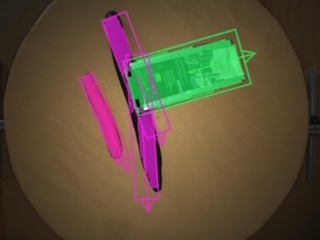}\\
			\includegraphics[width=0.23\textwidth]{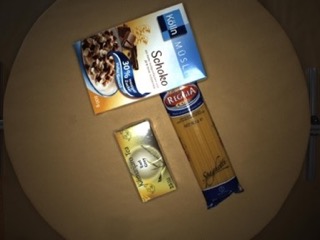}&
			\includegraphics[width=0.23\textwidth]{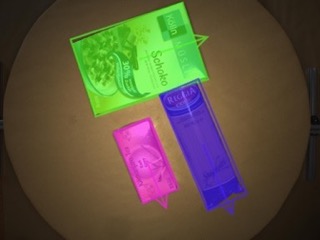}&
			\includegraphics[width=0.23\textwidth]{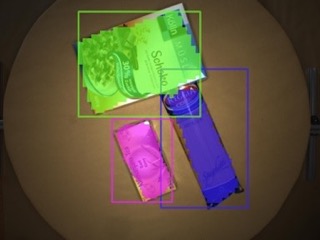}&
			\includegraphics[width=0.23\textwidth]{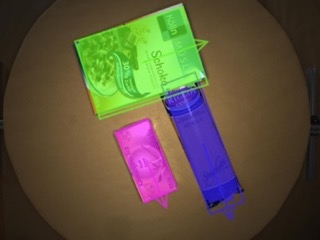}\\
			\includegraphics[width=0.23\textwidth]{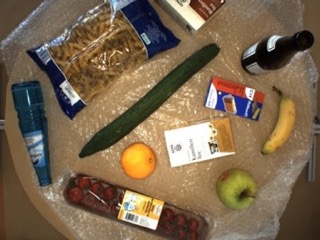}&
			\includegraphics[width=0.23\textwidth]{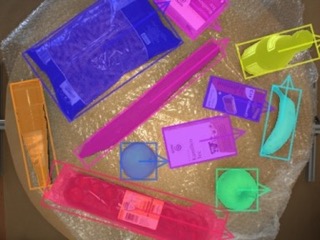}&
			\includegraphics[width=0.23\textwidth]{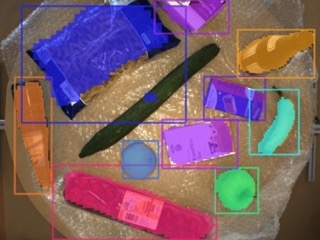}&
			\includegraphics[width=0.23\textwidth]{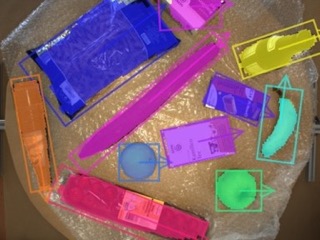}
		\end{tabular}
	\end{center}
	\caption{\textbf{\emph{D2S} results.} \emph{(from left to right)} input image, ground truth masks and generated ground truth oriented boxes, results based on axis-aligned boxes (\emph{AAIS}), results based on oriented boxes (\emph{OIS}) (best viewed digitally and with zoom)}
	\label{fig:ImgsD2S}
\end{figure*}

\begin{figure*}
	\begin{center}
		\setlength\tabcolsep{2pt}
		\begin{tabular}{c c c c}
			input & GT box and mask & \emph{AAIS bfm} & \emph{OIS} \\
			% FIRST row
			\includegraphics[width=0.23\textwidth]{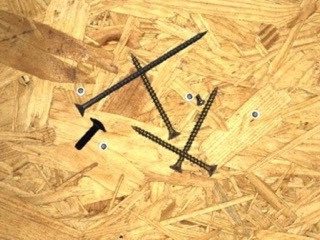}&
			\includegraphics[width=0.23\textwidth]{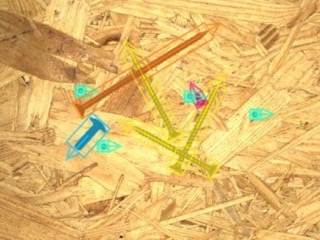}&
			\includegraphics[width=0.23\textwidth]{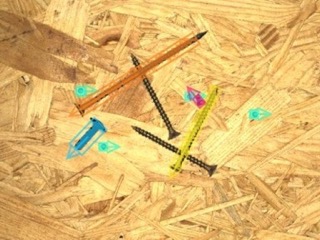}&
			\includegraphics[width=0.23\textwidth]{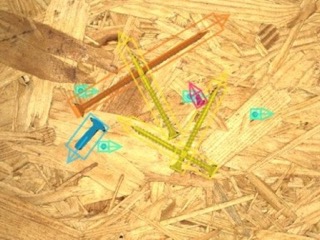}
			\\
			\includegraphics[width=0.23\textwidth]{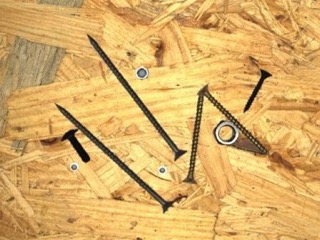}&
			\includegraphics[width=0.23\textwidth]{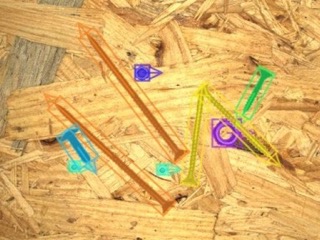}&
			\includegraphics[width=0.23\textwidth]{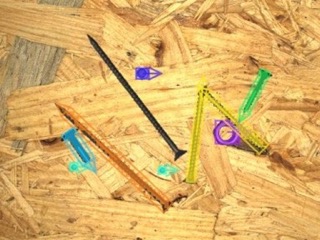}&
			\includegraphics[width=0.23\textwidth]{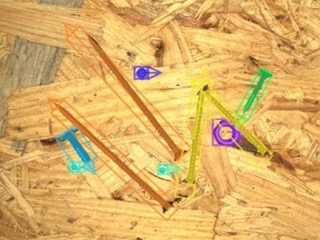}\\
			\includegraphics[width=0.23\textwidth]{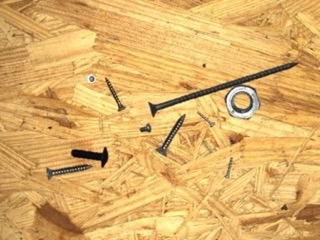}&
			\includegraphics[width=0.23\textwidth]{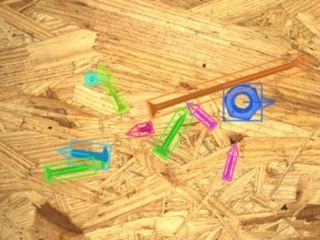}&
			\includegraphics[width=0.23\textwidth]{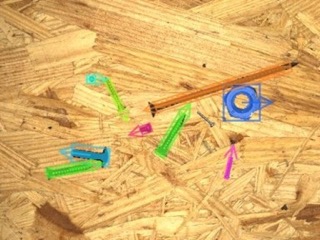}&
			\includegraphics[width=0.23\textwidth]{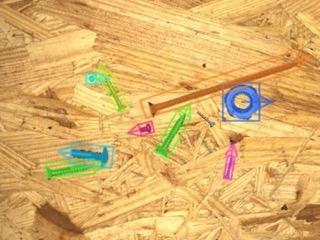}\\
			\includegraphics[width=0.23\textwidth]{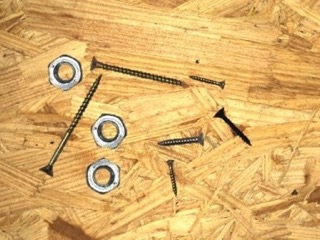}&
			\includegraphics[width=0.23\textwidth]{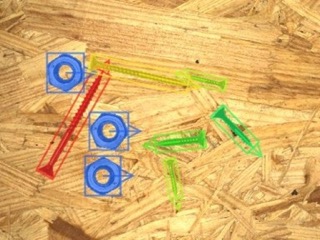}&
			\includegraphics[width=0.23\textwidth]{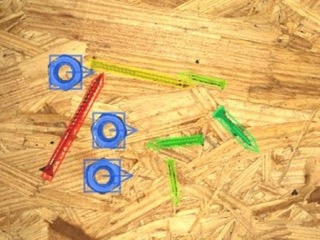}&
			\includegraphics[width=0.23\textwidth]{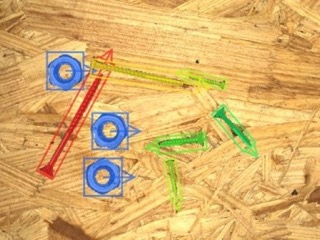}\\
			\includegraphics[width=0.23\textwidth]{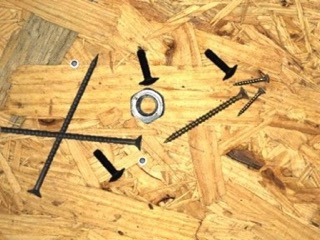}&
			\includegraphics[width=0.23\textwidth]{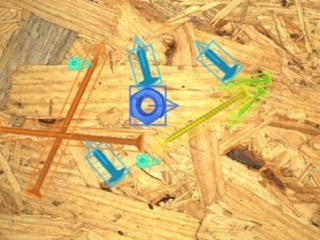}&
			\includegraphics[width=0.23\textwidth]{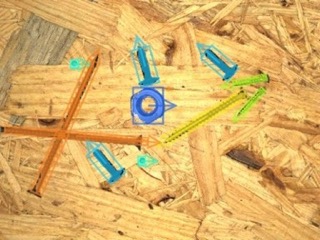}&
			\includegraphics[width=0.23\textwidth]{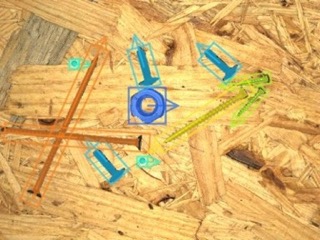}\\
			\includegraphics[width=0.23\textwidth]{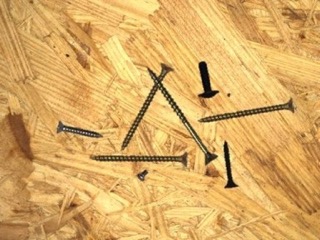}&
			\includegraphics[width=0.23\textwidth]{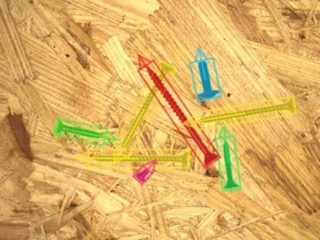}&
			\includegraphics[width=0.23\textwidth]{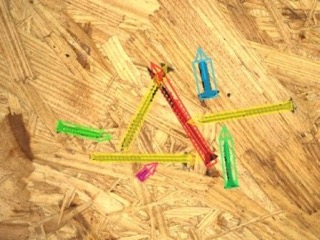}&
			\includegraphics[width=0.23\textwidth]{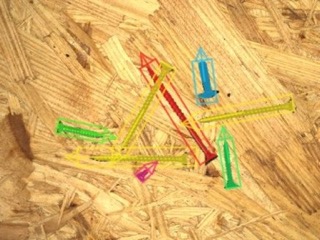}\\
			\includegraphics[width=0.23\textwidth]{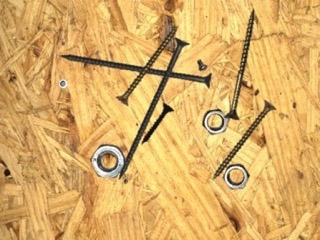}&
			\includegraphics[width=0.23\textwidth]{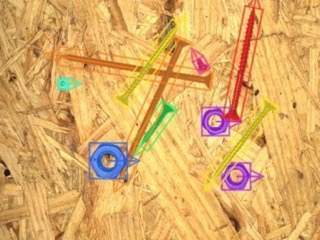}&
			\includegraphics[width=0.23\textwidth]{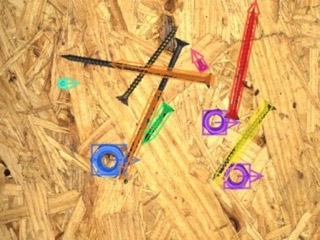}&
			\includegraphics[width=0.23\textwidth]{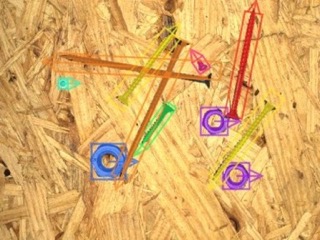}
		\end{tabular}
	\end{center}
	\caption{\textbf{\emph{Screws} (generated) results.} (best viewed digitally and with zoom)}
	\label{fig:ImgsScrewsAug}
\end{figure*}

\begin{figure*}
	\begin{center}
		\setlength\tabcolsep{2pt}
		\begin{tabular}{c c c c c c}
			input & GT box & \emph{AAIS bfm} & \emph{OIS} & \emph{OIS bfm} & \emph{OIS ofm} \\
			% FIRST row
			\includegraphics[width=0.16\textwidth]{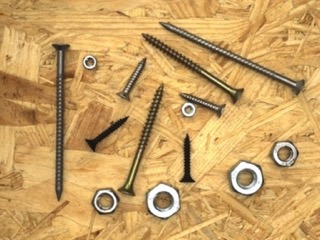}&
			\includegraphics[width=0.16\textwidth]{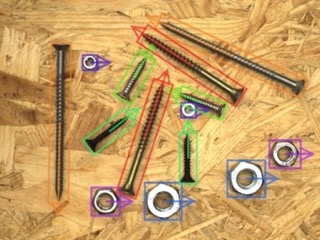}&
			\includegraphics[width=0.16\textwidth]{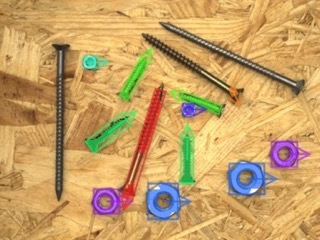}&
			\includegraphics[width=0.16\textwidth]{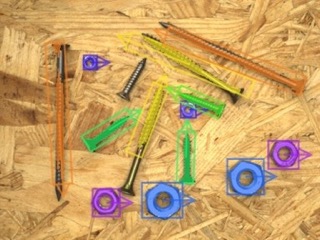}&
			\includegraphics[width=0.16\textwidth]{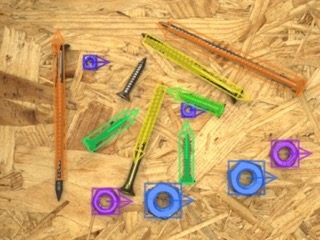}&
			\includegraphics[width=0.16\textwidth]{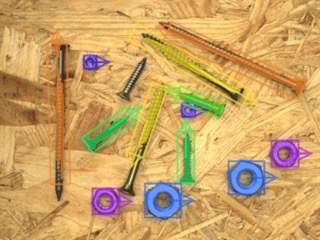}
			\\
			\includegraphics[width=0.16\textwidth]{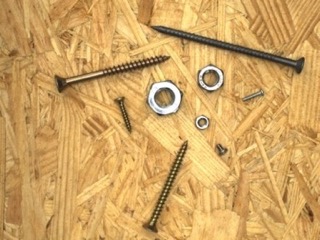}&
			\includegraphics[width=0.16\textwidth]{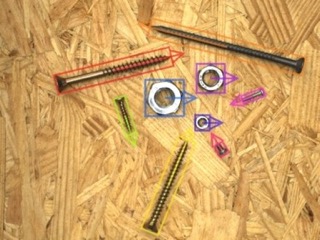}&
			\includegraphics[width=0.16\textwidth]{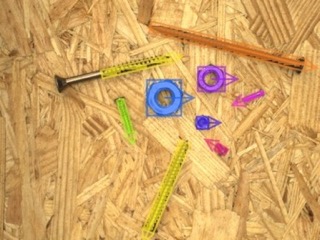}&
			\includegraphics[width=0.16\textwidth]{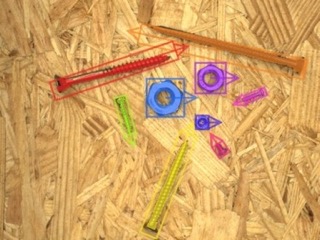}&
			\includegraphics[width=0.16\textwidth]{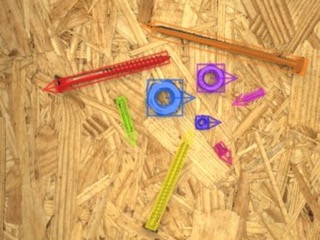}&
			\includegraphics[width=0.16\textwidth]{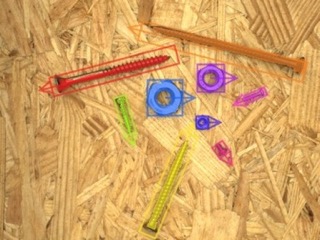}
			\\
			\includegraphics[width=0.16\textwidth]{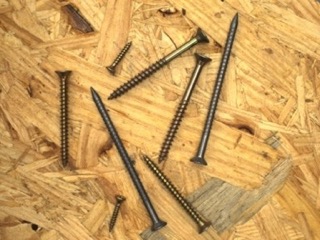}&
			\includegraphics[width=0.16\textwidth]{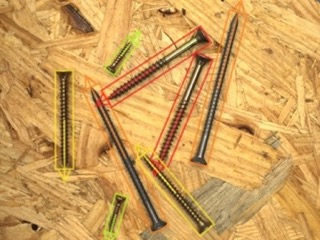}&
			\includegraphics[width=0.16\textwidth]{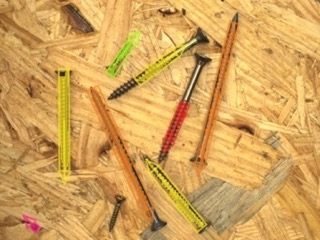}&
			\includegraphics[width=0.16\textwidth]{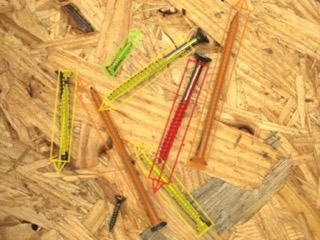}&
			\includegraphics[width=0.16\textwidth]{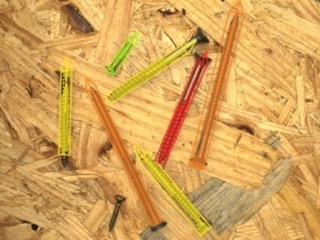}&
			\includegraphics[width=0.16\textwidth]{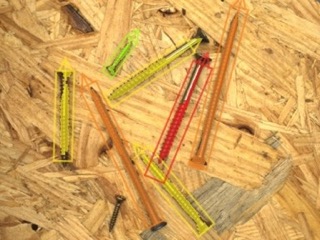}
			\\
			\includegraphics[width=0.16\textwidth]{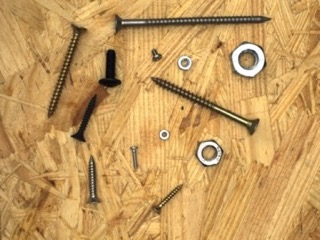}&
			\includegraphics[width=0.16\textwidth]{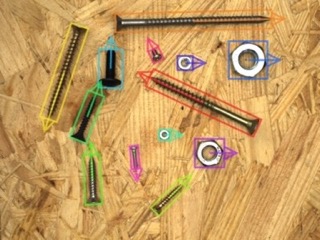}&
			\includegraphics[width=0.16\textwidth]{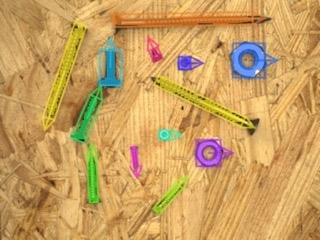}&
			\includegraphics[width=0.16\textwidth]{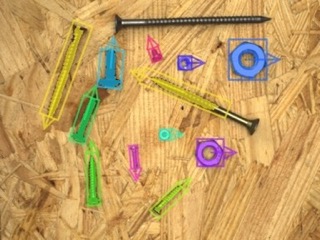}&
			\includegraphics[width=0.16\textwidth]{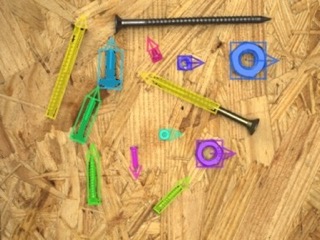}&
			\includegraphics[width=0.16\textwidth]{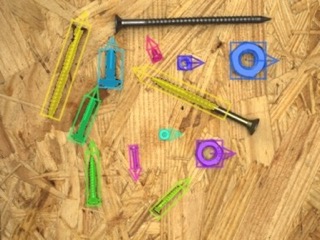}
			\\
			\includegraphics[width=0.16\textwidth]{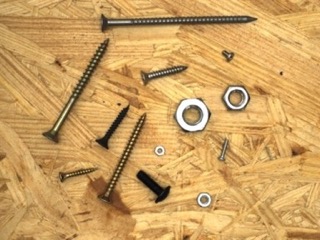}&
			\includegraphics[width=0.16\textwidth]{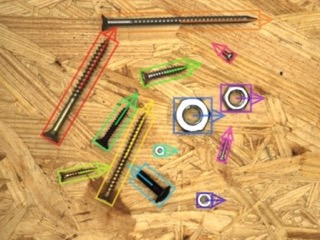}&
			\includegraphics[width=0.16\textwidth]{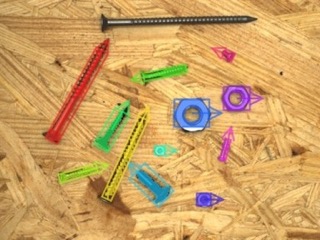}&
			\includegraphics[width=0.16\textwidth]{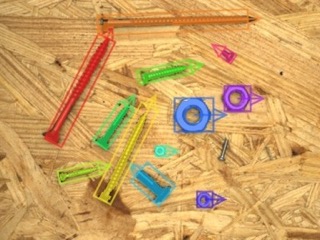}&
			\includegraphics[width=0.16\textwidth]{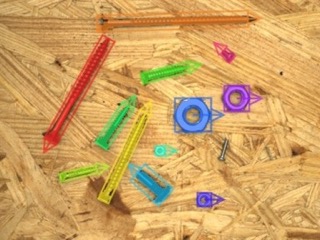}&
			\includegraphics[width=0.16\textwidth]{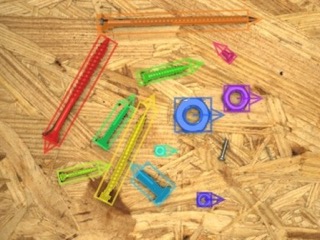}
			\\
			\includegraphics[width=0.16\textwidth]{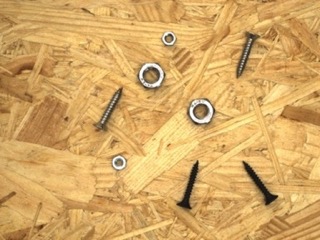}&
			\includegraphics[width=0.16\textwidth]{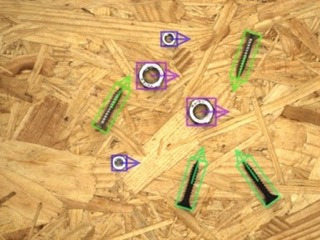}&
			\includegraphics[width=0.16\textwidth]{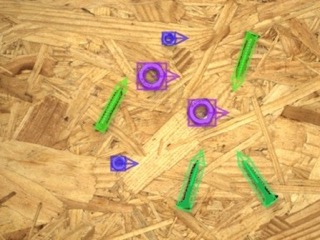}&
			\includegraphics[width=0.16\textwidth]{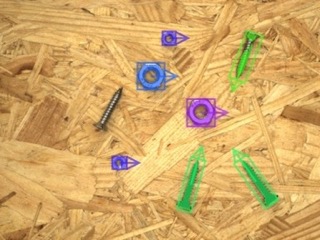}&
			\includegraphics[width=0.16\textwidth]{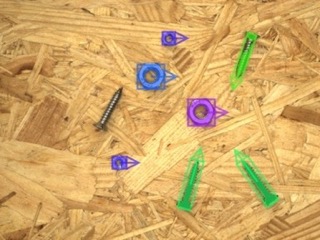}&
			\includegraphics[width=0.16\textwidth]{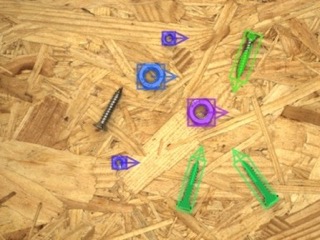}
			\\
			\includegraphics[width=0.16\textwidth]{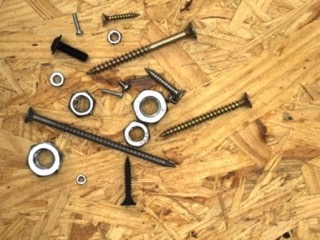}&
			\includegraphics[width=0.16\textwidth]{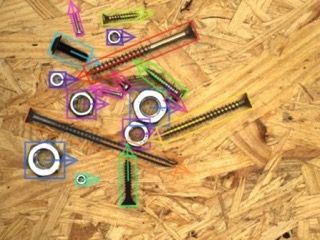}&
			\includegraphics[width=0.16\textwidth]{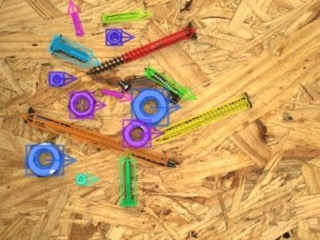}&
			\includegraphics[width=0.16\textwidth]{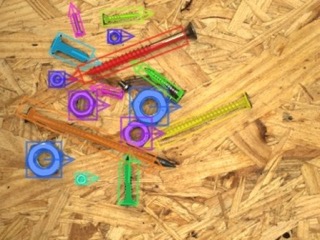}&
			\includegraphics[width=0.16\textwidth]{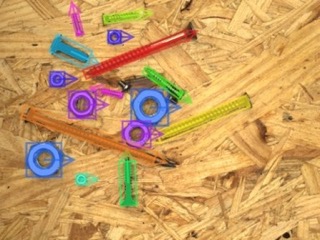}&
			\includegraphics[width=0.16\textwidth]{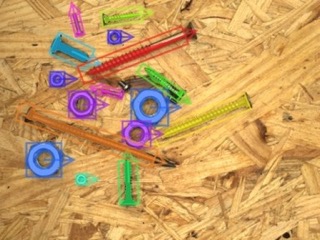}
			\\
			\includegraphics[width=0.16\textwidth]{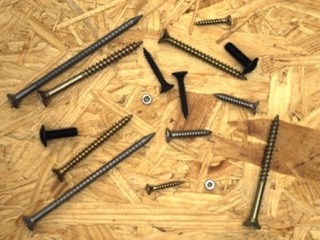}&
			\includegraphics[width=0.16\textwidth]{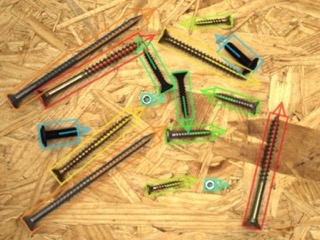}&
			\includegraphics[width=0.16\textwidth]{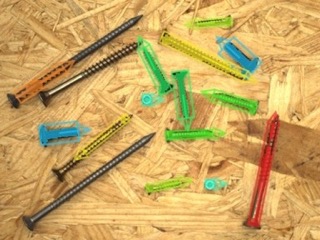}&
			\includegraphics[width=0.16\textwidth]{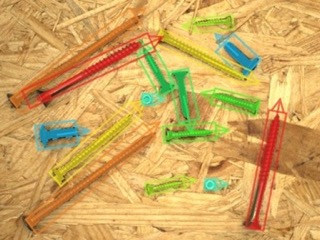}&
			\includegraphics[width=0.16\textwidth]{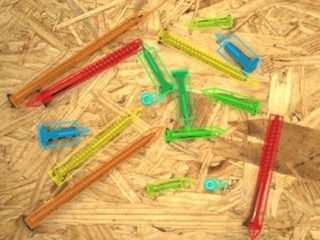}&
			\includegraphics[width=0.16\textwidth]{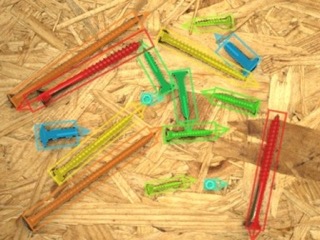}
			\\
			\includegraphics[width=0.16\textwidth]{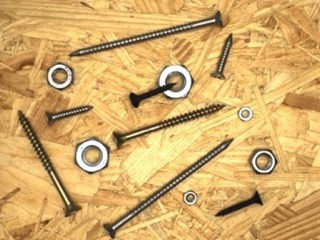}&
			\includegraphics[width=0.16\textwidth]{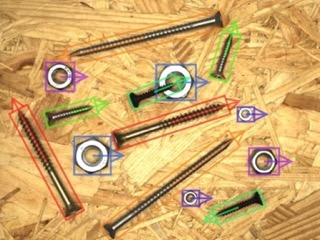}&
			\includegraphics[width=0.16\textwidth]{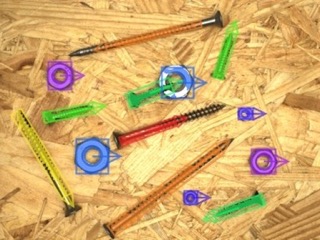}&
			\includegraphics[width=0.16\textwidth]{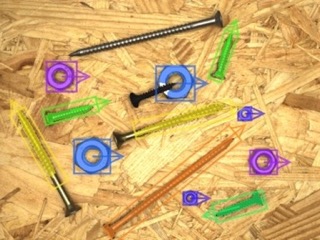}&
			\includegraphics[width=0.16\textwidth]{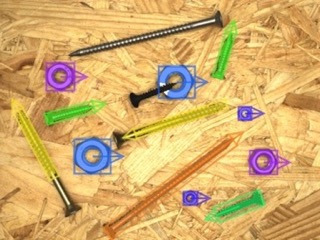}&
			\includegraphics[width=0.16\textwidth]{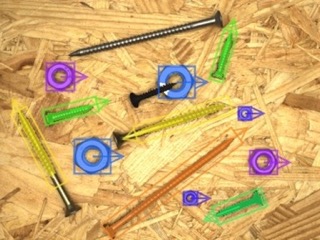}
		\end{tabular}
	\end{center}
	\vspace{0.3cm}
	\caption{\textbf{\emph{Screws} (real) results.} For \emph{OIS}, \emph{OIS bfm}, and \emph{OIS ofm} only the resulting box differs. (best viewed digitally and with zoom)}
	\label{fig:ImgsScrews}
\end{figure*}

\begin{figure*}
	\begin{center}
		\setlength\tabcolsep{2pt}
		\begin{tabular}{c c c c}
			input & GT box and mask & \emph{AAIS} & \emph{OIS} \\
			% FIRST row
			\includegraphics[width=0.23\textwidth]{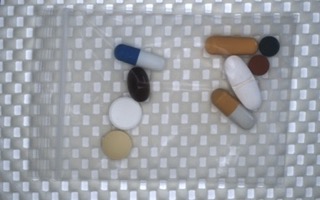}&
			\includegraphics[width=0.23\textwidth]{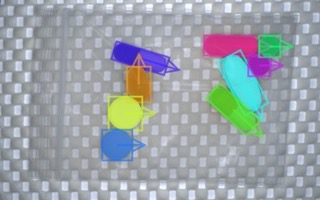}&
			\includegraphics[width=0.23\textwidth]{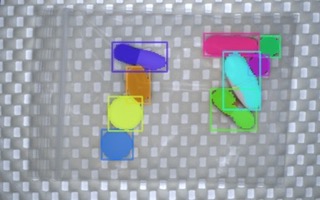}&
			\includegraphics[width=0.23\textwidth]{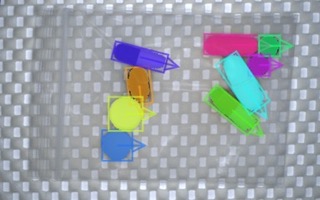}
			\\
			\includegraphics[width=0.23\textwidth]{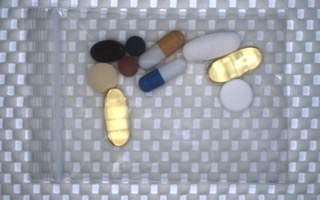}&
			\includegraphics[width=0.23\textwidth]{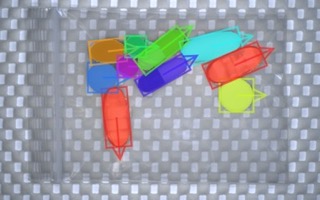}&
			\includegraphics[width=0.23\textwidth]{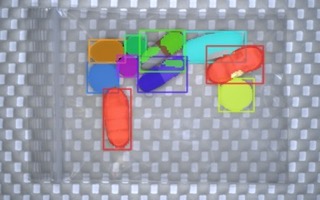}&
			\includegraphics[width=0.23\textwidth]{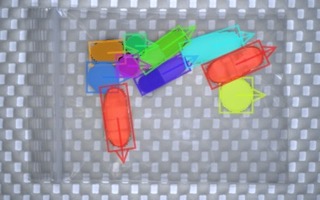}\\
			\includegraphics[width=0.23\textwidth]{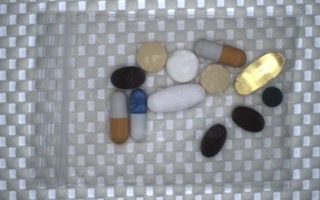}&
			\includegraphics[width=0.23\textwidth]{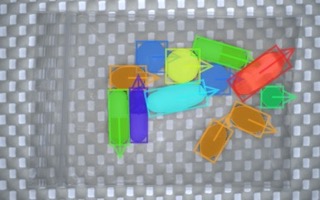}&
			\includegraphics[width=0.23\textwidth]{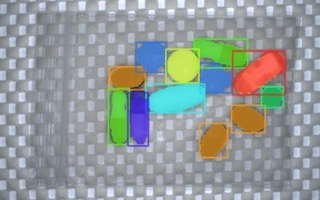}&
			\includegraphics[width=0.23\textwidth]{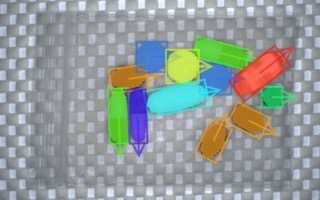}\\
			\includegraphics[width=0.23\textwidth]{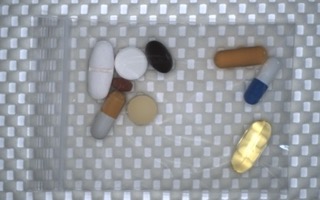}&
			\includegraphics[width=0.23\textwidth]{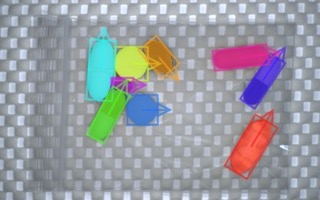}&
			\includegraphics[width=0.23\textwidth]{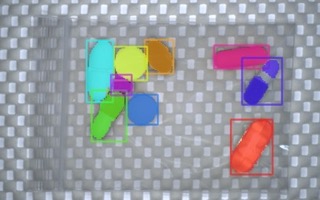}&
			\includegraphics[width=0.23\textwidth]{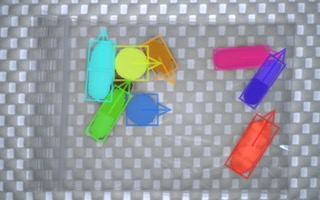}\\
			\includegraphics[width=0.23\textwidth]{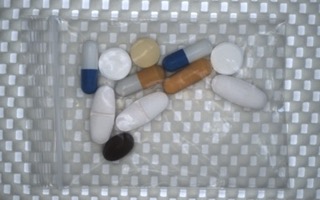}&
			\includegraphics[width=0.23\textwidth]{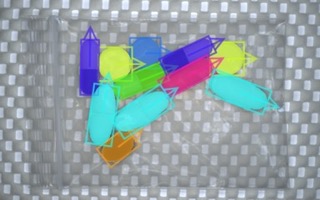}&
			\includegraphics[width=0.23\textwidth]{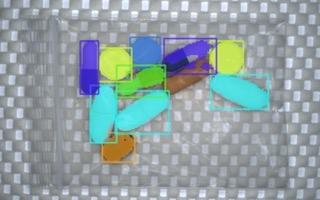}&
			\includegraphics[width=0.23\textwidth]{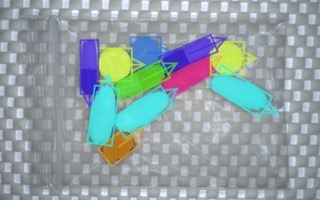}\\
			\includegraphics[width=0.23\textwidth]{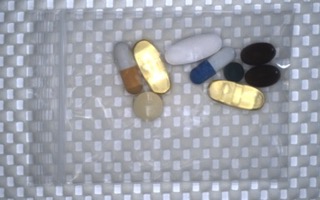}&
			\includegraphics[width=0.23\textwidth]{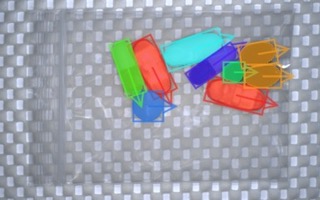}&
			\includegraphics[width=0.23\textwidth]{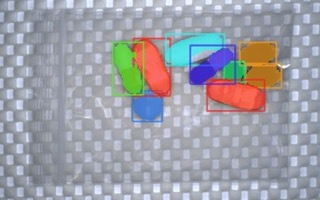}&
			\includegraphics[width=0.23\textwidth]{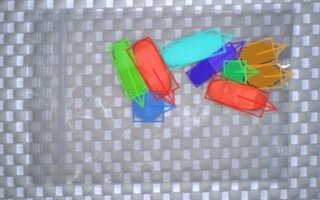}\\
			\includegraphics[width=0.23\textwidth]{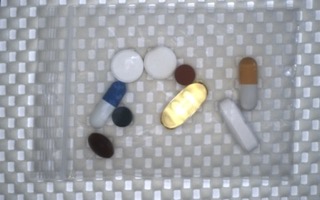}&
			\includegraphics[width=0.23\textwidth]{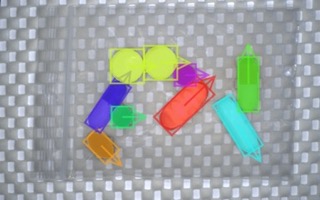}&
			\includegraphics[width=0.23\textwidth]{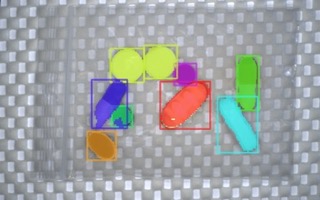}&
			\includegraphics[width=0.23\textwidth]{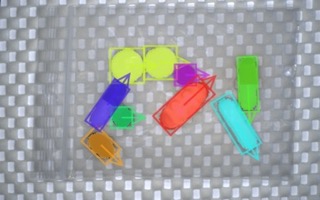}
		\end{tabular}
	\end{center}
	\caption{\textbf{\emph{Pill Bags} results.} Only 10 percent of the instance masks have been available during training. The shown images are from the test set (best viewed digitally and with zoom)}
	\label{fig:ImgsPillBags}
\end{figure*}

In this section we show some more qualitative results. \figref{fig:ImgsD2S} shows further results for \emph{D2S} and \figref{fig:ImgsScrewsAug} for \emph{Screws} (generated), respectively. In \figref{fig:ImgsScrews} further results on \emph{Screws} (real) are displayed. Here, in particular, the difference of \emph{OIS}, \emph{OIS box from mask (bfm)}, and \emph{OIS or from mask (ofm)} is visualized. Moreover, \figref{fig:ImgsPillBags} shows some more examples for \emph{Pill Bags} trained only with 10 percent of the instance masks. For all datasets, the displayed images are from the test set and were not seen during training.

The results also contain failure cases. Typical failure cases for \emph{OIS} result from inaccurate box predictions, e.g. the orientation $(\phi)$, the center $(r, c)$, or semi-axes $(l1, l2)$ are not predicted correctly. %Moreover, on \emph{Screws}, the model has difficulties with very small objects. We believe this problem could be fixed by finding more suitable anchor and FPN parameters.

%%%%%%%%%%%%%%%%% D2S Results Figure %%%%%%%%%%%%%%%%%%%
%%%%%%%%%%%%%%%%% D2S Results Figure %%%%%%%%%%%%%%%%%%%

%%%%%%%%%%%%%%%%% Screws Aug Results Figure %%%%%%%%%%%%%%%%%%%
%%%%%%%%%%%%%%%%% Screws Aug Results Figure %%%%%%%%%%%%%%%%%%%

%%%%%%%%%%%%%%%%% Screws Results Figure %%%%%%%%%%%%%%%%%%%
%%%%%%%%%%%%%%%%% Screws Results Figure %%%%%%%%%%%%%%%%%%%

%%%%%%%%%%%%%%%%% Pill Bags Results Figure %%%%%%%%%%%%%%%%%%%
%%%%%%%%%%%%%%%%% Pill Bags Results Figure %%%%%%%%%%%%%%%%%%%

\section{Calculation of orientation}

In this section, we show how we calculate the orientation of a region as used in \mbox{Section 4.2}.

Consider a region $R$ consisting of $n$ pixels with row coordinates $r_i$ and column coordinates $c_i$, $i=1,\dots, n$. We denote the center of gravity by $g = (r_0, c_0)$:
\[ (r_0, c_0) = \frac{1}{n} \sum_{i=1}^n (r_i, c_i) \]
Further, the second moments of $R$, $M_{ij}$ are given by:
\[ M_{ij} = \sum_{(r,c) \in R} (r_0 - r)^i (c_0 - c)^j \]
We calculate the orientation $\phi$ of $R$ by fitting an ellipse to the region that has the same aspect ratio and orientation and get:
\[ \phi = -\frac{1}{2} \text{atan2}(2M_{11}, M_{02} - M_{20}) \]

\begin{figure}[h]
	\centering
	\includegraphics[width=0.25\textwidth]{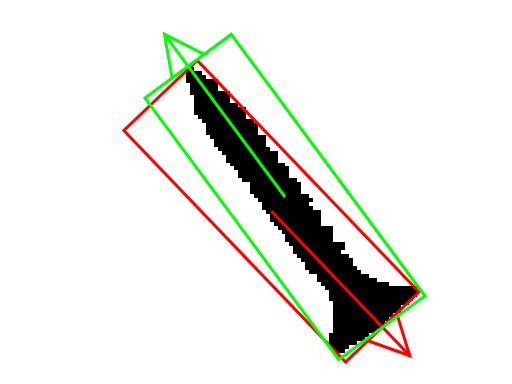}
	\caption{{\bf Example for orientation of region.} (black) underlying region $R$ of a screw, (red) smallest oriented bounding box. (green) bounding box with orientation obtained from the region's orientation. It points from the screws head to its tail and is therefore more consistent over different screw instances}
	\label{fig:ExampleOrientation}
\end{figure}

An example where this method to extract the orientation has benefits is given in \figref{fig:ExampleOrientation}. To obtain the bounding box with orientation $\phi$ calculated as above, we rotate the mask to be axis-aligned, get the axis-aligned bounding box, and rotate it back to have orientation $\phi$.

%%%%%%%%%%%%%%%%% Example Orientation Figure %%%%%%%%%%%%%%%%%%%
%%%%%%%%%%%%%%%%% Example Orientation Figure %%%%%%%%%%%%%%%%%%%

\section{Detailed model configurations}

As described in the main paper, we use two different architectures based on Mask RCNN \cite{he_2017_mask} (M) and RetinaMask \cite{fu2019retinamask} (R). In contrast to the original implementation of RetinaMask we do not use the self-adjusting smooth L1-loss. % where the first stage is an RPN with fg/bg and box heads, the second stage are the Faster RCNN \cite{ren_2015_faster} class and box heads, and the third stage is a mask head as in \mbox{Mask RCNN}.

In comparison to \cite{ren_2015_faster}, in the RPN we use branches similar to a RetinaNet with focal loss \cite{lin2017focal}. Therefore, we do not need box sampling during training. Moreover, we do not share weights between the fg/bg branch and the box branch and also do not share weights between different FPN-levels.

In the final mask prediction we use both a class specific and a class agnostic NMS. For both the $\text{IoU}$ threshold can be set individually.

To make our results reproducible, we show all shared model settings for Mask RCNN with explanations in \tabref{tab:ParamDefault1} and \tabref{tab:ParamDefault2}. In the following, we share the parameters that were changed for the application of our architecture on \emph{D2S} and \emph{Screws}. However, note that we do not claim that the given parameters are tuned optimally.

\subsection{D2S}

In comparison to \emph{AAIS}, for \emph{OIS} only the anchor aspect ratios and angles are adapted (\emph{AAIS} has no anchor angles), see \tabref{tab:D2SModelParams}. For RetinaMask \cite{fu2019retinamask} (R), we adapted the anchor aspect ratios for \emph{OIS} such that in \emph{AAIS} and \emph{OIS} the same number of anchors is used.

\begin{table}
	\centering
	\small
	\begin{tabular}{l | c | c } \hline
		parameter    & \emph{AAIS} & \emph{OIS} \\ \hline \hline
		\multicolumn{3}{c}{anchor parameter} \\ \hline
		aspect ratios & (0.5, 1.0, 2.0) (M) & (0.4, 0.7, 1.0) (M) \\
		
		\multirow{2} {*}{} & (0.3, 0.6, 0.9, & \multirow{2}{*}{(0.4, 0.9) (R)} \\
		& 1.11, 1.66, 3.33) (R) & \\
		%& (0.3, 0.6, 0.9, 1.11, 1.66, 3.33)  (R) & (0.4, 0.9) (R) \\
		angles & - (M, R) & ($-\frac{\pi}{3}, 0.0, \frac{\pi}{3}$) (M, R) \\ \hline
		\multicolumn{3}{c}{RPN parameter} \\ \hline
		num convolutions & \multicolumn{2}{c}{3 (M, R)}  \\
		max num post NMS & \multicolumn{2}{c}{50 (R)}  \\ \hline
		\multicolumn{3}{c}{RCNN head parameter (M)} \\ \hline
		max num post NMS & \multicolumn{2}{c}{28} \\ \hline
		\multicolumn{3}{c}{Mask head parameter} \\ \hline
		num convs & \multicolumn{2}{c}{2} \\ \hline
	\end{tabular}
	\caption{{\bf \emph{D2S} detailed model parameter settings.} Parameters for Mask RCNN (M) and RetinaMask (R) are shown. See text for further information}
	\label{tab:D2SModelParams}
\end{table}

\subsection{Screws}

In comparison to \emph{D2S}, for \emph{Screws}, the main differences are the lighter backbone and that $\text{IgnoreDirection}$ is set to $\text{false}$, such that orientations in the range $(-\pi, \pi)$ are predicted. Accordingly, anchors for the full orientation range are used. We adapt the RPN parameters for high recall. All changed parameters for Mask RCNN are shown in \tabref{tab:ScrewsModelParams}.

\begin{table}
	\centering
	\small
	\begin{tabular}{l | c | c } \hline
		parameter    & \emph{AAIS} & \emph{OIS} \\ \hline \hline
		\multicolumn{3}{c}{backbone parameter} \\ \hline
		backbone  & \multicolumn{2}{c}{SqueezeNet \cite{iandola2016squeezenet}} \\
		dataset pretraining & \multicolumn{2}{c}{OpenImages \cite{OpenImages}} \\ 
		freeze at & \multicolumn{2}{c}{0 (no freeze)} \\ \hline
		\multicolumn{3}{c}{anchor parameter} \\ \hline
		aspect ratios & (0.5, 1.0, 2.0) & (0.15, 0.35, 0.65, 1.0) \\
		\multirow{2} {*}{angles} & \multirow{2}{*}{-} & ($-\frac{2\pi}{3}$, $-\frac{\pi}{3}$, $0.0$, \\
		& & $\frac{\pi}{3}$, $-\frac{2\pi}{3}$, $\pi$) \\
		IgnoreDirection & - & false \\ \hline
		\multicolumn{3}{c}{RPN parameter} \\ \hline
		min level   & \multicolumn{2}{c}{2} \\
		max level   & \multicolumn{2}{c}{4} \\
		max num pre NMS & \multicolumn{2}{c}{400} \\
		max num post NMS & \multicolumn{2}{c}{512} \\
		NMS IoU threshold & \multicolumn{2}{c}{0.9} \\
		conv dim & \multicolumn{2}{c}{256} \\
		final conv kernel size & \multicolumn{2}{c}{$3\times3$} \\ \hline
		\multicolumn{3}{c}{RPN training parameter} \\ \hline
		box loss weight & \multicolumn{2}{c}{2.0} \\ \hline
		\multicolumn{3}{c}{RCNN head parameter} \\ \hline
		RoI min level & \multicolumn{2}{c}{2} \\
		RoI max level & \multicolumn{2}{c}{4} \\
		RoI canonical level & \multicolumn{2}{c}{4} \\
		NMS IoU threshold & \multicolumn{2}{c}{1.0 (off)} \\
		NMS IoU threshold agn & \multicolumn{2}{c}{0.2} \\
		max num pre NMS & \multicolumn{2}{c}{512} \\ \hline
		\multicolumn{3}{c}{RCNN head training parameter} \\ \hline
		batch size per img & \multicolumn{2}{c}{512} \\
		ratio num fg & \multicolumn{2}{c}{0.9} \\ 
		box angle weight & \multicolumn{2}{c}{2.0} \\ \hline
		\multicolumn{3}{c}{Mask head parameter} \\ \hline
		mask min score & \multicolumn{2}{c}{0.4} \\ \hline
	\end{tabular}
	\caption{{\bf \emph{Screws} detailed Mask RCNN model parameter settings.} See text for further information}
	\label{tab:ScrewsModelParams}
\end{table}

\subsection{Pill Bags}

The detailed model parameter settings for RetinaMask trained on \emph{Pill Bags} are shown in \tabref{tab:PillBagsModelParams}. The min and max level, anchor aspect ratios and angles (only for \emph{OIS}) as well as the NMS IoU threshold and NMS IoU threshold agn were calculated automatically based on the distribution of bounding boxes in the training set. For the NMS IoU thresholds we require a minimum value of 0.1.

\begin{table}
	\centering
	\small
	\begin{tabular}{l | c | c } \hline
		parameter    & \emph{AAIS} & \emph{OIS} \\ \hline \hline
		\multicolumn{3}{c}{backbone parameter} \\ \hline
		image size & \multicolumn{2}{c}{512 x 320} \\
		backbone & \multicolumn{2}{c}{SqueezeNet \cite{iandola2016squeezenet}} \\
		dataset pretraining & \multicolumn{2}{c}{OpenImages \cite{OpenImages}} \\ 
		freeze at & \multicolumn{2}{c}{0 (no freeze)} \\ \hline
		\multicolumn{3}{c}{anchor parameter} \\ \hline
		num subscales & 3 & 2 \\
		\multirow{2}{*}{aspect ratios} & (0.54, 1.00, & \multirow{2}{*}{(0.38, 0.65, 1.00)} \\
		& 1.52, 2.31) & \\
		\multirow{2}{*}{angles} & \multirow{2}{*}{-} & (-0.80, 0.0, \\
		& &  0.75, 1.55) \\ \hline
		\multicolumn{3}{c}{RPN parameter} \\ \hline
		min level   & 2 & 3 \\ 
		\cline{2-3}
		max level   & \multicolumn{2}{c}{4} \\
		max num pre NMS & \multicolumn{2}{c}{500} \\
		max num post NMS & \multicolumn{2}{c}{30} \\
		\cline{2-3}
		NMS IoU threshold & 0.14 & 0.1 \\
		NMS IoU threshold agn & 0.25 & 0.1 \\
		\cline{2-3}
		final conv kernel size & \multicolumn{2}{c}{$3\times3$} \\ \hline
		\multicolumn{3}{c}{RPN training parameter} \\ \hline
		fgNegThresh & \multicolumn{2}{c}{0.4} \\ \hline
		\multicolumn{3}{c}{RCNN head parameter} \\ \hline
		RoI min level & \multicolumn{2}{c}{2} \\
		RoI max level & \multicolumn{2}{c}{4} \\
		RoI canonical level & \multicolumn{2}{c}{4} \\
		NMS IoU threshold & \multicolumn{2}{c}{1.0 (off)} \\
		NMS IoU threshold agn & \multicolumn{2}{c}{0.2} \\
		max num pre NMS & \multicolumn{2}{c}{512} \\ \hline
		\multicolumn{3}{c}{Mask head parameter} \\ \hline
		num convs & \multicolumn{2}{c}{2} \\ \hline
		\multicolumn{3}{c}{Mask head training parameter} \\ \hline
		mask loss weight & \multicolumn{2}{c}{2.0} \\ \hline
	\end{tabular}
	\caption{{\bf \emph{Pill Bags} detailed RetinaMask model parameter settings.} See text for further information}
	\label{tab:PillBagsModelParams}
\end{table}

\section{Solver settings}

All general solver settings are depicted in \tabref{tab:SolverSettings}. As for the model parameters, we do not claim that these parameters are optimal. However, we found that they lead to a stable training and we show them in such detail to make our results reproducible.

\begin{table}
	\centering
	\small
	\begin{tabular}{l | c | c | c} \hline
		parameter    & \emph{Screws} & \emph{D2S} & \emph{Pill Bags}\\ \hline \hline
		
		momentum   & \multicolumn{3}{c}{0.9} \\
		\cline{2-4}
		weight decay & \multicolumn{2}{c|}{1e-6} & 1e-5 \\
		\cline{2-4}
		initial learning rate & \multicolumn{3}{c}{0.001} \\
		lr strategy & \multicolumn{3}{c}{step} \\
		gamma (lr factor) & \multicolumn{3}{c}{(0.1 0.01)} \\ \hline
		steps & (30, 50) & (30, 40) (M) & (15, 30)\\
		& & 25 (R) & \\
		max epochs & 60 & 50 (M) & 35 \\
		& & 40 (R) & \\
		batch size & 2 & 4 (M) & 2\\
		& & 2 (R) & \\
		
		warmup factor & 0.33 & 0.1 & 0.33 \\
		warmup iterations & 200 & 500 & 200\\ \hline
		
	\end{tabular}
	\caption{{\bf Solver parameters}}
	\label{tab:SolverSettings}
\end{table}

As we train all layers inside the heads and the FPN from scratch, we found that the training was diverging easily. This problem could be fixed by a dynamic loss weighting strategy as described in the following subsection.

\subsection{Dynamic loss weights}

Besides the backbone, all layers within the FPN, RPN, RCNN and mask branches are not pretrained but initialized randomly. Both the RPN branches as well as the RCNN heads are using features from the FPN. However, in the beginning of the training, the box deltas and classes that are predicted by the RCNN heads are mostly wrong and generate large losses. This results in large gradients that are not aligned with the gradients of the RPN. This leads to frequent divergence of the training. 

Therefore, at the beginning of the training, when the RPN loss is still high, we reduce both the box loss weight and the class loss weight to low initial values ($ilw_{box}$ and $ilw_{class}$). During training, we dynamically increase those loss weights linearly to the final box loss weight $flw_{box}$ and the final class loss weight $flw_{cls}$. This is done over $n_{dlw}$ iterations: whenever the running average RPN loss $\bar{l}_{RPN}$ falls below the current RPN loss threshold $Tl_{RPN}$, the box and class loss weights are increased and $Tl_{RPN}$ is multiplied by a factor $\delta Tl_{RPN} < 1$.

More formally: Initialize the current loss weights for box $lw_{box} = ilw_{box}$ and class $ lw_{cls} = ilw_{cls}$. Initialize the current RPN loss threshold $Tl_{RPN} = iTl_{RPN}$. Initialize the current iteration $i_{dlw} = 0$. Let $\bar{l}_{RPN}$ be the total RPN loss averaged over the last 10 iterations. Let $\delta lw_{box}$ and $\delta lw_{cls}$ be the stepsizes for the box and class loss weights, respectively:
\begin{align*} 
\delta lw_{box} &= \frac{flw_{box} - ilw_{box}}{n_{dlw}} \\
\delta lw_{cls} &= \frac{flw_{cls} - ilw_{cls}}{n_{dlw}}
\end{align*}
Now always if $\bar{l}_{RPN} < Tl_{RPN}$ and $i_{dlw} < n_{dlw}$, we update:
\begin{align*}
lw_{box} &\leftarrow lw_{box} + \delta lw_{box} \\
lw_{cls} &\leftarrow lw_{cls} + \delta lw_{cls} \\
Tl_{RPN} &\leftarrow Tl_{RPN} \cdot \delta Tl_{RPN} \\
i_{dlw} &\leftarrow i_{dlw} + 1
\end{align*}

For our experiments we use the parameters as shown in \tabref{tab:DLWParams}. They remain unchanged for \emph{AAIS} and \emph{OIS}. The dynamic loss weight strategy is only used for Mask RCNN \cite{he_2017_mask}, but not for RetinaMask \cite{fu2019retinamask}.

\begin{table}
	\centering
	\small
	\begin{tabular}{l | c | c } \hline
		parameter    & \emph{Screws} & \emph{D2S} \\ \hline \hline
		
		$ilw_{box}$ & \multicolumn{2}{c}{0.01} \\
		$ilw_{cls}$ & \multicolumn{2}{c}{0.01} \\
		$flw_{box}$ & \multicolumn{2}{c}{1.0} \\ \hline
		$flw_{cls}$ & 1.0 & 2.0 \\ \hline
		$iTl_{RPN}$ & \multicolumn{2}{c}{0.5} \\
		$n_{dlw}$ & \multicolumn{2}{c}{50} \\ \hline
	\end{tabular}
	\caption{{\bf Dynamic loss weight parameters}}
	\label{tab:DLWParams}
\end{table}

\begin{table*}
	\centering
	\small
	\begin{tabular}{l | c | l } \hline
		parameter    & default value & explanation \\ \hline \hline
		\multicolumn{3}{c}{backbone parameter} \\ \hline
		image size & $512\times384$ & input image dimensions (w, h) \\
		backbone  & ResNet50 \cite{He_2016_CVPR} & backbone classifier network \\
		dataset pretraining & ImageNet \cite{deng_2009_imagenet} & dataset for pretraining \\ 
		freeze at & 3 & The first n layers of the backbone are frozen \\ \hline
		\multicolumn{3}{c}{RPN parameter} \\ \hline
		min level   & 3 & minimum used FPN level \\
		max level   & 5 & maximum used FPN level \\
		max num pre NMS & 512 & maximum number of boxes before NMS \\
		&        & is applied (per FPN level) \\
		max num post NMS & 256 & maximum number of boxes after NMS  \\
		& & was applied (total)\\
		NMS IoU threshold & 0.8 & IoU threshold for NMS \\
		min side length & 1.0 & minimum side length of output boxes \\
		num convolutions & 4 & number of convolutions in RPN branches \\
		conv dim & 128 & number of kernels for the intermediate \\ & &convolutions of the RPN \\
		final conv kernel size & $1\times1$ & kernel size of final convolution in RPN branches \\
		min score & 0.05 & minimal score of output boxes (only inference) \\ \hline
		\multicolumn{3}{c}{anchor parameter} \\ \hline
		num subscales & 3 & number of subscales/octaves for each anchor \\
		aspect ratios & (0.5, 1.0, 2.0) & aspect ratios of anchors \\
		angles & ($-\frac{2\pi}{3}, 0.0, \frac{2\pi}{3}$) & orientations of anchors (only for \emph{OIS}) \\ 
		IgnoreDirection & true & See paper, only for oriented boxes \\ \hline
		\multicolumn{3}{c}{RPN training parameter} \\ \hline
		$\text{fgPosThresh}$ & 0.5 & positive threshold for anchor assignment \\
		$\text{fgNegThresh}$ & 0.3 & negative threshold for anchor assignment \\
		\emph{swb2bg} & false & if true, anchors below $\text{fgNegThresh}$ are always \\ & &assigned to background \\
		fg/bg loss weight & 1.0 & loss weight for fg/bg branch \\
		box loss weight & 1.0 & loss weight for box branch \\
		box center weight & 2.0 & loss factor for (r,c) prediction \\
		box dim weight & 1.0 & loss factor for $(w,h)/(l1,l2)$ prediction \\
		box angle weight & 1.0 & loss factor for $\phi$ prediction \\
		focal loss gamma & 2.0 & gamma in RPN focal loss \\
		focal loss beta & 0.11 & beta in RPN focal loss \\ \hline
	\end{tabular}
	\caption{{\bf Model parameters with explanations part 1}}
	\label{tab:ParamDefault1}
\end{table*}

\begin{table*}[!ht]
	\centering
	\small
	\begin{tabular}{l | c | l } \hline
		parameter    & default value & explanation \\ \hline \hline
		\multicolumn{3}{c}{RCNN head parameter} \\ \hline
		RoI min level$^*$ & 3 & minimum FPN level for RoI pooling \\
		RoI max level$^*$ & 5 & maximum FPN level for RoI pooling \\
		RoI canonical scale$^*$ & 224 & anchors of canonical scale are distributed to \\
		RoI canonical level$^*$ & 5 & the canonical level see \cite{ren_2015_faster} \\
		RoI pool mode$^*$ & roi\_pool & In this work, we only use the original RoI \\ & &pooling (no RoIAlign) \\
		RoI grid size (M)& (7, 7) & RoI pooling grid size (w, h) for RCNN head \\
		MLP head dim (M)& 256 & dim of intermediate fc layer of RCNN \\ &&head MLP \\
		NMS IoU threshold & 0.3 & class specific IoU threshold (only inference)\\
		NMS IoU threshold agn & 0.3 & class agnostic IoU threshold (only inference) \\
		max num pre NMS & 256 & maximum number of boxes before NMS\\& &is applied (only inference) \\
		max num post NMS & 30 & maximum number of boxes after NMS\\ & &was applied (only inference) \\
		min score & 0.5 & minimal score of output boxes (only inference) \\ \hline
		\multicolumn{3}{c}{RCNN head training parameter} \\ \hline
		$\text{fgPosThresh}$ & 0.7 & positive threshold for box assignment \\
		$\text{fgNegThresh}$ & 0.5 & negative threshold for box assignment \\
		\emph{swb2bg} & true & if true, boxes below $\text{fgNegThresh}$ are always\\ & &assigned to background \\
		batch size per img & 256 & maximum number of samples per image \\
		ratio num fg & 0.75 & Ratio of fg/bg samples in batch \\
		class loss weight & dlw & we use a dynamic loss weight strategy \\
		box loss weight & dlw & we use a dynamic loss weight strategy \\
		box center weight & 2.0 & loss factor for (r,c) prediction \\
		box dim weight & 1.0 & loss factor for $(w,h)/(l1,l2)$ prediction \\
		box angle weight & 1.0 & loss factor for $\phi$ prediction \\
		\hline
		\multicolumn{3}{c}{Mask head parameter} \\ \hline
		num convs & 4 & number of convolutions in mask head \\
		conv dim & 128 & number of kernels in intermediate convs \\
		RoI grid size & (14, 14) & RoI pooling output size (w, h) for mask head \\
		Mask final grid size & (28, 28) & mask output and target size (w, h) \\
		mask min score & 0.5 & minimum score for mask prediction (only inf.) \\ \hline    
		\multicolumn{3}{c}{Mask head training parameter} \\ \hline
		$\text{fgPosThresh}$ & 0.7 & positive threshold for final box assignment \\
		$\text{fgNegThresh}$ & 0.6 & negative threshold for final box assignment \\
		\emph{swb2bg} & true & if true, boxes below $\text{fgNegThresh}$ are always \\ &&assigned to background \\
		mask loss weight & 1.0 & weight of mask branch loss \\ \hline
	\end{tabular}
	\caption{{\bf Model parameters with explanations part 2.} ($^*$) Parameters are shared between RCNN and mask heads}
	\label{tab:ParamDefault2}
\end{table*}

\end{document}